%% file: main.tex
\documentclass[runningheads]{llncs}
\usepackage[T1]{fontenc}
\usepackage{graphicx}
\usepackage{amsmath,amsfonts,amssymb}
\usepackage{xcolor}
\usepackage{adjustbox}
\usepackage{colortbl}
\usepackage{tabularx}
\usepackage{booktabs} 
\usepackage{graphicx}
\usepackage{subfig}
\usepackage[linesnumbered, vlined, ruled]{algorithm2e}
\SetArgSty{textup}
\usepackage{enumerate}
\usepackage{prettyref}
\usepackage{color, colortbl}
\usepackage{braket}
\definecolor{MyHiLiRow}{gray}{0.9}
\usepackage{tcolorbox}
\usepackage{caption}
\usepackage{seqsplit}
\usepackage{bm}
\usepackage{nameref}
\usepackage{multirow}
\usepackage{xr}
\usepackage{tikz}
\usepackage{enumerate}

\usepackage{mathtools}
\usepackage[hidelinks]{hyperref}

\usepackage[whole]{bxcjkjatype}

\def\vec#1{\mathbf{#1}}
\def\set#1{\mathcal{#1}}

\newcommand\tabblue[1]{{\color[HTML]{1f77b4}{#1}}}

\newcommand\tabred[1]{{\color[HTML]{d62728}{#1}}}

\definecolor{refpt}{RGB}{44,160,44}   
\definecolor{nearpt}{RGB}{255,127,14} 
\definecolor{solpt}{RGB}{31,119,180} 

\newcommand{\pref}[1]{\prettyref{#1}}
\newrefformat{fig}{Fig.~\ref{#1}}
\newrefformat{supfig}{Figure~\ref{#1}}
\newrefformat{tab}{Table~\ref{#1}}
\newrefformat{suptab}{Figure~\ref{#1}}
\newrefformat{sec}{Section~\ref{#1}}
\newrefformat{subsec}{Section~\ref{#1}}
\newrefformat{app}{Appendix~\ref{#1}}
\newrefformat{alg}{Algorithm~\ref{#1}}
\newrefformat{property}{Property~\ref{#1}}
\newrefformat{theorem}{Theorem~\ref{#1}}
\newrefformat{corollary}{Corollary~\ref{#1}}
\newrefformat{proposition}{Proposition~\ref{#1}}
\newrefformat{def}{Definition~\ref{#1}}
\newrefformat{eq}{equation~(\ref{#1})}

\begin{document}
\title{Quantitative Performance Analysis of\\Stopping Criteria for CMA-ES}
%
%
\author{Ryoji Tanabe\orcidID{0000-0003-4049-0393}}
%
%
\institute{Yokohama National University, Yokohama, Japan\\
\email{tanabe-ryoji-sn@ynu.ac.jp}}
\maketitle              

\input{abstract}

\input{introduction}
\input{preliminary_cma}

\input{preliminary_criteria}
\input{setup}

\input{results}
\input{conclusion}

\subsubsection*{Acknowledgments.}

This work was supported by JSPS KAKENHI Grant Number \seqsplit{25K03194}.

\bibliographystyle{splncs04}
\bibliography{reference}

\end{document}

%% file: abstract.tex
\begin{abstract}

Covariance matrix adaptation evolution strategy (CMA-ES) is a state-of-the-art black-box optimization algorithm.
In general, CMA-ES uses a portfolio of multiple stopping criteria to automatically determine when to stop the search.
This mechanism aims to avoid unnecessary consumption of the function evaluation budget during stagnation.
Stopping criteria play an important role in CMA-ES, particularly when restart strategies are employed.
However, the effectiveness of stopping criteria in CMA-ES remains poorly understood.
To address this issue, this paper investigates how the 11 stopping criteria in CMA-ES behave on the noiseless BBOB function set.
The performance of the stopping criteria is quantitatively evaluated based on the optimal stopping point in terms of the number of function evaluations in a single run of CMA-ES.
Our results show that, although which stopping criterion is triggered first depends significantly on the sample size $\lambda$ and the dimension $n$, \texttt{tolflatfitness} and \texttt{tolfun} are frequently the first criteria to be triggered among the portfolio of 11 stopping criteria.
We also demonstrate that \texttt{tolfunhist} and the portfolio achieve the highest stopping accuracy in most cases.
In addition, our results show that the \texttt{tolfun} and \texttt{tolfunhist} criteria are frequently triggered before CMA-ES reaches complete stagnation.



\keywords{CMA-ES \and stopping criteria \and empirical analysis}
\end{abstract}

%% file: introduction.tex
\section{Introduction}
\label{sec:introduction}

\noindent \textit{General context.}
This paper considers minimization of a black-box objective function $f: \mathbb{R}^n \to \mathbb{R}$.
In the black-box setting, the structure of $f$ is unknown a priori, and only the objective value of a candidate solution is available for the search.

Covariance matrix adaptation evolution strategy (CMA-ES)~\cite{Hansen16a,HansenK04,HansenO01} is one of the state-of-the-art black-box optimization algorithms.
For each iteration, CMA-ES samples a set of $\lambda$ offspring from an $n$-dimensional multivariate normal distribution $\mathcal{N}(\vec{m}, \sigma^2 \vec{C})$, where $\vec{m} \in \mathbb{R}^n$ is a mean vector, $\sigma \in \mathbb{R}$ is a step-size, and $\vec{C} \in \mathbb{R}^{n \times n}$ is a covariance matrix.
CMA-ES adaptively adjusts the three distribution parameters $\vec{m}$, $\sigma$, and $\vec{C}$ using the best $\mu$ individuals among the $\lambda$ offspring according to their objective values.
CMA-ES has desirable invariance properties~\cite{Hansen16a}, including the rotational invariance and the invariance to order preserving transformations of the objective value.


Any evolutionary algorithm has at least one stopping criterion~\cite{LiuZZ18,SafeCPB04}, including CMA-ES.
Ideally, the search of an evolutionary algorithm should be stopped when further improvement of the best-so-far solution is no longer expected.
Otherwise, the function evaluation budget is wasted.
Budget-based stopping criteria are the most commonly used in the evolutionary computation community.
In this approach, the search is stopped when the number of function evaluations exceeds the maximum number of function evaluations ($\texttt{FE}^{\mathrm{max}}$) or the number of iterations exceeds the maximum number of iterations ($t^{\mathrm{max}}$).
A main drawback of budget-based stopping criteria is the difficulty in selecting suitable values for $\texttt{FE}^{\mathrm{max}}$ and $t^{\mathrm{max}}$ a priori.
For this reason, evolutionary algorithms require stopping criteria that automatically determine whether to stop the search~\cite{LiuZZ18,SafeCPB04}.

Stopping criteria in CMA-ES have been developed incrementally.
To the best of our knowledge, non-budget-based stopping criteria in CMA-ES were first discussed in~\cite{HansenK04} in 2004, where three stopping criteria were introduced: \texttt{tolfun}, \texttt{tolx}, and \texttt{tolconditioncov}.
In ~\cite{AugerH05a,AugerH05}, Auger and Hansen benchmarked two restart versions of CMA-ES on the CEC2005 function set, including IPOP-CMA-ES.
They introduced the following three stopping criteria: \texttt{tolfunhist}, \texttt{noeffectaxis}, and \texttt{noeffectcoord}.
In \cite{Hansen09}, Hansen introduced two additional stopping criteria for BIPOP-CMA-ES: \texttt{tolupsigma} and \texttt{tolstagnation}.
A tutorial paper of CMA-ES~\cite{Hansen16a} summarizes eight stopping criteria proposed in the literature.
%
The latest version of the Python library \texttt{pycma}~\cite{Hansen19} implements 15 stopping criteria, including 12 non-budget-based ones.


In addition to budget-based stopping criteria, CMA-ES generally employs multiple non-budget-based stopping criteria.
For example, IPOP-CMA-ES~\cite{AugerH05} and BIPOP-CMA-ES~\cite{Hansen09} use six and nine non-budget-based stopping criteria, respectively. 
When at least one stopping criterion is satisfied during the search, the CMA-ES run is stopped.
Stopping criteria play an important role, especially in restart versions of CMA-ES~\cite{NobelVKSB24}, including IPOP-CMA-ES and BIPOP-CMA-ES.
This is because stopping criteria determine when a restart is triggered and thus directly influence the performance of restart versions of CMA-ES.
Stopping criteria also significantly influence the performance of sequential algorithm portfolios~\cite{Fukunaga98,HubermanLH97,Schapermeier25}, which run multiple optimization algorithms sequentially.



\vspace{0.5em}
\noindent \textit{Motivation.}
However, the effectiveness of stopping criteria remains poorly understood.
Although the 12 non-budget-based stopping criteria are available in \texttt{pycma}, it is unknown how they contribute to stopping the search.
Some previous studies (e.g., ~\cite{Lopez-IbanezLS12,SmitE10}) tuned the hyperparameters of IPOP-CMA-ES, including those related to stopping criteria.
For example, the threshold values of \texttt{tolfun}, \texttt{tolx}, and \texttt{tolfunhist} were tuned by \texttt{irace}~\cite{LopezIbanez16} in~\cite{Lopez-IbanezLS12}.
Unfortunately, these studies focused on the performance of IPOP-CMA-ES and did not analyze the stopping criteria.
Some previous studies (e.g., \cite{Nobel0B21,RijnDB18}) investigated the effectiveness of each  component in CMA-ES, but they did not consider stopping criteria.

\vspace{0.5em}
\noindent \textit{Contribution.}
Motivated by the above discussion, this paper investigates how the stopping criteria in CMA-ES work on the noiseless BBOB function set (\texttt{bbob})~\cite{hansen2012fun}.
We quantitatively evaluate the performance of the stopping criteria using the optimal number of function evaluations required to stop a single run of CMA-ES, which is a recently proposed concept in~\cite{KitamuraT26}.
This is the first study to systematically investigate the effectiveness of stopping criteria in CMA-ES.


\vspace{0.5em}
\noindent \textit{Terminology.}
The terms ``stopping criteria$"$ and ``termination criteria$"$ have been used interchangeably in the evolutionary computation community. For example, both terms are used in~\cite{Hansen16a}. For consistency, we use ``stopping criteria$"$ throughout this paper.

\vspace{0.5em}
\noindent \textit{Outline.}
Sections \ref{sec:cmaes} and \ref{sec:review_sc} describe CMA-ES and its stopping criteria, respectively.
%
\pref{sec:setup} describes our experimental setup.
\pref{sec:results} shows the results of the analysis. 
\pref{sec:conclusion} concludes this paper.

\vspace{0.5em}
\noindent \textit{Code and data availability.}
The code used in this work is available at \url{https://gist.github.com/ryojitanabe/626b163ce245a3b851629985138c98ca}.

%% file: preliminary_cma.tex

\section{CMA-ES}
\label{sec:cmaes}

Below, we briefly describe CMA-ES~\cite{Hansen16a,HansenK04,HansenO01}.
At the beginning of the search, CMA-ES initializes the three internal parameters: the mean vector $\vec{m} \in \mathbb{R}^n$, the step-size $\sigma \in \mathbb{R}$, and the covariance matrix $\vec{C} \in \mathbb{R}^{n \times n}$.
In addition, two evolution paths $\vec{p}_{\text{c}} \in \mathbb{R}^n$ and $\vec{p}_{\sigma} \in \mathbb{R}^{n}$ are also initialized as zero vectors.
Here, $\vec{p}_{\text{c}}$ is used for the rank-one update of $\vec{C}$, and $\vec{p}_{\sigma}$ is used for cumulative step-size adaptation of $\sigma$.
After the initialization, the following procedure is repeated until a stopping  condition is met.

For each iteration $t$, first, CMA-ES samples $\lambda$ offspring $\vec{x}^{(t)}_{1}, \dots, \vec{x}^{(t)}_{\lambda}$ from the multivariate normal distribution $\mathcal{N}(\vec{m}^{(t)}, (\sigma^{(t)})^2 \vec{C}^{(t)})$.
Here, the mean vector $\vec{m}^{(t)}$ is the center of the sampling distribution.
The step-size $\sigma^{(t)}$ controls the mutation strength.
The covariance matrix $\vec{C}$ determines the shape of the sampling distribution.

Next, the $\lambda$ offspring $\vec{x}^{(t)}_{1}, \dots, \vec{x}^{(t)}_{\lambda}$ are sorted according to their objective values such that $f(\vec{x}^{(t)}_{1}) \leq  \dots  \leq f(\vec{x}^{(t)}_{\lambda})$.
Then, the five parameters $\vec{m}$, $\sigma$, $\vec{C}$, $\vec{p}_{\text{c}}$, and $\vec{p}_{\sigma}$ are updated for the next iteration $t+1$ based on the best $\mu$ of the $\lambda$ offspring\footnote{Active covariance matrix adaptation~\cite{JastrebskiA06} uses all $\lambda$ offspring to update $\vec{C}$.}, where $\mu \leq \lambda$.
The new mean vector $\vec{m}^{(t+1)}$ is set to a weighted average of the best $\mu$ offspring.
The two evolution paths $\vec{p}_{\text{c}}^{(t+1)}$ and $\vec{p}_{\sigma}^{(t+1)}$ are updated using $\vec{C}^{(t)}$, $\vec{m}^{(t+1)}$, and $\vec{m}^{(t)}$.
Then, the step-size $\sigma^{(t+1)}$ is updated based on $\vec{p}_{\sigma}^{(t+1)}$.
Finally, the new covariance matrix $\vec{C}^{(t+1)}$ is obtained by rank-$\mu$ and rank-one updates based on $\vec{C}^{(t)}$ and $\vec{p}_{\text{c}}^{(t+1)}$.

%% file: preliminary_criteria.tex
\begin{table}[t]
\centering
\begin{minipage}[t]{0.59\linewidth}
  \centering
  \footnotesize
  \caption{Classification of stopping criteria.} 
  \label{tab:classif}
  \begin{tabular}{llcc}
    \toprule
    Group & Stopping criteria\\
    \midrule
Budget & \texttt{maxiter}\\
& \texttt{maxfevals}\\
 & \texttt{timeout}\\
    \midrule
Fitness &\texttt{tolfun}\\
&\texttt{tolfunrel} (not used)\\
&\texttt{tolfunhist}\\
&\texttt{tolflatfitness}\\
&\texttt{tolstagnation}\\
    \midrule
Convergence&\texttt{tolxstagnation}\\
& \texttt{tolx}\\ 
& \texttt{noeffectcoord}\\ 
& \texttt{noeffectaxis}\\ 
\midrule
Divergence &\texttt{tolconditioncov}\\
&\texttt{tolfacupx}\\
&\texttt{tolupsigma}\\
    \bottomrule
  \end{tabular}
\end{minipage}
\begin{minipage}[t]{0.4\linewidth}
  \centering
  \footnotesize
  \caption{Default settings.}
  \label{tab:sc_parameters}
  \begin{tabular}{llcc}
    \toprule
    Parameters & Values\\
    \midrule
$\epsilon_{\texttt{tfun}}$ & $10^{-11}$\\
$\epsilon_\texttt{tfunr}$ & $0$\\
$\epsilon_\texttt{thist}$ & $10^{-12}$\\
$T_{\texttt{tflat}}$ & $1$\\
$\epsilon_\texttt{tcond}$ & $10^{14}$\\
$\epsilon_\texttt{tfacupx}$ & $10^3$\\
$\epsilon_\texttt{tsigma}$ & $10^{20}$\\
\raisebox{0.5em}{$T_{\texttt{tstag}}$} & \shortstack{$\lfloor 100+100$\\$\times n \times 1.5 / \mu \rfloor$}\\
$\epsilon_{\texttt{txstag}}$ & $10^{-9}$\\
$\epsilon_{\texttt{tx}}$ & $10^{-11}$\\ 
    \bottomrule
  \end{tabular}
\end{minipage}
\end{table}


\section{Stopping criteria in CMA-ES}
\label{sec:review_sc}


As described in \pref{sec:introduction}, multiple stopping criteria are generally used in CMA-ES.
However, there is no standardized choice of stopping criteria.
This work considers the stopping criteria in \texttt{pycma}~\cite{Hansen19}, which is one of the most popular CMA-ES libraries and is maintained by the original author of CMA-ES.


\pref{tab:classif} shows the 15 stopping criteria implemented in \texttt{pycma}. 
\pref{tab:sc_parameters} also shows their default hyperparameter values.
We categorize the 15 stopping criteria into four groups: budget-based, fitness-based, convergence-based, and divergence-based ones.
The three budget-based criteria are commonly used in the evolutionary computation community.
As described in \pref{sec:introduction}, \texttt{maxiter} and \texttt{maxfevals} are based on $t^{\mathrm{max}}$ and $\texttt{FE}^{\mathrm{max}}$, respectively.
The \texttt{timeout} criterion stops the search when the elapsed time exceeds a predefined limit, but this criterion is seldom used in black-box optimization.
The five fitness-based criteria determine whether to stop the search based solely on fitness values.
In contrast, the four convergence-based and three divergence-based criteria use internal parameters of CMA-ES, such as the mean vector $\vec{m}$, step-size $\sigma$, and covariance matrix $\vec{C}$.
On the one hand, convergence-based criteria determine whether internal parameters have converged to particular values.
For example, \texttt{tolxstagnation} aims to detect the convergence of $\vec{m}$.
On the other hand, the divergence-based criteria determine whether internal parameters diverge.
For example, \texttt{tolfacupx} stops the search when $\sigma$ and $\vec{C}$ grow excessively large.
Below, we describe the 12 non-budget-based stopping criteria.


\subsection{Fitness-based stopping criteria}



\noindent \textbf{$\bullet$} \texttt{tolfun}
In the current iteration $t$, the minimum and maximum objective values among the $\lambda$ offspring are obtained as follows: $F_{\text{min}}^{(t)} = \min_{i \in \{1, \dots, \lambda \}} f (\vec{x}^{(t)}_i) $ and $F_{\text{max}}^{(t)} = \max_{i \in \{1, \dots, \lambda \}} f (\vec{x}^{(t)}_i)$.
Let $\set{H}^{(t)} = \{F_{\text{min}}^{(t)}, \dots, F_{\text{min}}^{(t-l)}\}$ be a set of minimum objective values over the most recent $l$ iterations, where $l=10+ \lfloor 30n/\lambda \rfloor$.
The search is stopped when \textit{both} of the following conditions are met: (i) $F_{\text{max}}^{(t)} - F_{\text{min}}^{(t)} < \epsilon_{\texttt{tfun}}$ and 
(ii) $\max \set{H}^{(t)} - \min \set{H}^{(t)} < \epsilon_{\texttt{tfun}}$.
%

\vspace{0.3em}
\noindent \textbf{$\bullet$} \texttt{tolfunrel}
Unlike \texttt{tolfun}, \texttt{tolfunrel} is based on the relative improvement of the objective value since the first iteration $t=1$.
Let $F_{\text{med}}^{(t)}$ be the median objective value among the $\lambda$ offspring at iteration $t$.
The search is stopped when the following condition is met: $F_{\text{max}}^{(t)} - F_{\text{min}}^{(t)} < \epsilon_{\texttt{tfunr}} (F_{\text{med}}^{(1)} - F_{\text{med}}^{(t)})$.
As shown in \pref{tab:sc_parameters}, the default value of $\epsilon_{\texttt{tfunr}}$ is zero.
Therefore, no stopping is triggered based on \texttt{tolfunrel} under the default setting of \texttt{pycma}.

\vspace{0.3em}
\noindent \textbf{$\bullet$} \texttt{tolfunhist}
This stops the search when the following condition is satisfied: $\max \set{H}^{(t)} - \min \set{H}^{(t)} < \epsilon_{\texttt{thist}}$.
Clearly, the \texttt{tolfunhist} criterion is equivalent to the second condition of the \texttt{tolfun} criterion when $\epsilon_{\texttt{thist}} = \epsilon_{\texttt{tolfun}}$.
Thus, the \texttt{tolfunhist} criterion measures the difference between the smallest and largest objective values in $\set{H}^{(t)}$.

\vspace{0.3em}
\noindent \textbf{$\bullet$} \texttt{tolflatfitness}
This criterion aims to detect stagnation on flat fitness plateaus.
%
Let $F_{\lfloor \alpha\lambda \rfloor}^{(t)}$ be the $\alpha$th percentile of the objective values among the $\lambda$ offspring at iteration $t$.
If $F_{\text{min}}^{(t)} = F_{\lfloor 0.75\lambda \rfloor}^{(t)}$, we say that the fitness distribution is flat.
The search is stopped when $F_{\text{min}}^{(t)} = F_{\lfloor 0.75\lambda \rfloor}^{(t)}$ for $T_{\texttt{tflat}}$ consecutive iterations.
Since the default value of $T_{\texttt{tflat}}$ in \texttt{pycma} is $1$, the \texttt{tolflatfitness} criterion stops the search immediately when $F_{\text{min}}^{(t)} = F_{\lfloor 0.75\lambda \rfloor}^{(t)}$.


\vspace{0.3em}
\noindent \textbf{$\bullet$} \texttt{tolstagnation}
%
Let $\set{L}$ and $\set{M}$ be sets of minimum and median objective values among the $\lambda$ offspring recorded every five iterations, respectively.
For example, when $t=20$, $\set{L}^{(t)} = \{F_{\text{min}}^{(5)}, F_{\text{min}}^{(10)},$ $F_{\text{min}}^{(15)}, $ $F_{\text{min}}^{(20)}\}$ and $\set{M}^{(t)} = \{F_{0.5\lambda}^{(5)}, F_{0.5\lambda}^{(10)}, F_{0.5\lambda}^{(15)}, F_{0.5\lambda}^{(20)}\}$.
When $t > 20\,000$, the oldest elements in $\set{L}$ and $\set{M}$ are removed every five iterations.
The search is stopped when \textit{all} of the following seven conditions are satisfied:
(a) $t \geq l$,
(b) $t > n \left(5 + \frac{100}{\lambda}\right)$, 
(c) $ \frac{\texttt{FE} - \texttt{FE}^{\text{last}}}{\lambda} > \frac{T_{\texttt{tstag}}}{2}$, 
(d) $|\set{L}^{(t)}| > 100 $, 
(e) $|\set{L}^{(t)}| > 2l$, 
(f) $\text{median} (\set{L}^{(1)}, $ $ \dots, $ $ \set{L}^{(t)}) \geq \text{median} (\set{L}^{(l)}, \dots, \set{L}^{(2l)})$,  and
(g) $\text{median} (\set{M}^{(1)}, $ $ \dots, $ $ \set{M}^{(t)}) \geq \text{median} (\set{M}^{(l)}, $ $\dots, $ $\set{M}^{(2l)})$.
%
%
Here, \texttt{FE} denotes the number of function evaluations.
$\texttt{FE}^{\text{last}}$ denotes the number of function evaluations at which the best-so-far objective value was last updated.
The function ``$\text{median}$" returns the median of its inputs.
The period length $l$ is defined as follows:
$l := \max \left\lfloor \left\{\frac{T_{\texttt{tstag}}}{5 \times 2}, \frac{|\set{L}^{(t)}|}{10} \right\} \right\rfloor$.
Conditions (a)--(e) determine when the \texttt{tolstagnation} criterion starts to apply.
Conditions (f)--(g) measure the improvement in the quality of the $\lambda$ offspring over the periods $1, \dots, t$ and $l, \dots, 2l$.






\subsection{Convergence-based stopping criteria}

\noindent \textbf{$\bullet$} \texttt{tolxstagnation}
In this criterion, the mean vector $\vec{m}$ is considered to have moved when $\|\vec{m}^{(t)} - \vec{m}^{(t^\text{last})} \| \geq \epsilon^{(t)}$.
Here, $t^\text{last}$ denotes the iteration at which the mean vector last moved.
At the beginning of the search, $t^\text{last}$ is initialized to $1$.
The threshold parameter $\epsilon^{(t)}$ for measuring the movement of $\vec{m}$ is calculated as follows:
\begin{align}
\epsilon^{(t)} &:= \epsilon_{\texttt{txstag}} \sqrt{\max \left\{1, \frac{t - t^{\text{last}}}{T} \right\} },\notag 
\end{align}
where $T = 20 + 0.1 t$.
The \texttt{tolxstagnation} criterion stops the search when the mean vector $\vec{m}$ does not move for more than $T$ iterations.
Formally, the search is stopped when $t - t^{\text{last}} > T$.






\vspace{0.3em}
\noindent \textbf{$\bullet$} \texttt{tolx}
This criterion measures the coordinate-wise mutation strengths of the normal distribution and the convergence of the evolution path $\vec{p}_{\text{c}}$.
For convenience in describing the \texttt{pycma} implementation, we introduce a diagonal matrix $\vec{S}^{(t)} = \mathrm{diag}(s_1^{(t)}, \dots, s_n^{(t)})$ corresponding to \texttt{sigma\_vec.scaling}.\footnote{This notation is introduced only for describing the \texttt{pycma} implementation.}
Let also $[\vec{C}^{(t)}]_{i,i}$ be the $i$-th diagonal element of $\mathbf{C}^{(t)}$.
The search is stopped when the following two conditions are satisfied for all $i \in \{1, \dots, n\}$:
$\sigma^{(t)} s_i^{(t)} \sqrt{[\vec{C}^{(t)}]_{i,i}} < \epsilon_{\texttt{tolx}}$
and
$\sigma^{(t)} s_i^{(t)} p_{\text{c},i}^{(t)} < \epsilon_{\texttt{tolx}}$.
Here, $\sigma^{(t)} s_i^{(t)} \sqrt{[\vec{C}^{(t)}]_{i,i}}$ represents the coordinate-wise mutation strength in the $i$-th coordinate.

\vspace{0.3em}
\noindent \textbf{$\bullet$} \texttt{noeffectcoord}
This criterion checks whether a coordinate-wise perturbation along an axis-aligned direction has no numerical effect on the current mean vector $\vec{m}^{(t)}$.
The search is stopped when the following condition is satisfied for at least one $i \in \{1, \dots, n\}$:
$m_i^{(t)} = m_i^{(t)} + 0.2 \, \sigma^{(t)} s_i^{(t)} \sqrt{[\vec{C}^{(t)}]_{i,i}}$.

%


\vspace{0.3em}
\noindent \textbf{$\bullet$} \texttt{noeffectaxis}
Unlike \texttt{noeffectcoord}, \texttt{noeffectaxis} checks whether small perturbations along the eigen-directions have no numerical effect on the current mean vector $\vec{m}^{(t)}$.
Let $\mathbf{C}^{(t)} = \mathbf{B}^{(t)} (\mathbf{D}^{(t)})^2 (\mathbf{B}^{(t)})^\top$ be the eigendecomposition of the covariance matrix $\mathbf{C}^{(t)}$.
Let $\vec{b}_i^{(t)}$ be the $i$-th column of $\mathbf{B}^{(t)}$.
Let also $d_i^{(t)}$ denote the $i$-th diagonal element of $\mathbf{D}^{(t)}$ for each $i \in \{1, \dots, n\}$.
Then, the search is stopped when the following condition is satisfied for all $i \in \{1, \dots, n\}$: $\vec{m}^{(t)} = \vec{m}^{(t)} + 0.1 \, \sigma^{(t)} d_i^{(t)} \left(\vec{S}^{(t)} \vec{b}_i^{(t)}\right)$.




\subsection{Divergence-based stopping criteria}

\vspace{0.3em}
\noindent \textbf{$\bullet$} \texttt{tolfacupx}
This criterion checks whether the coordinate-wise mutation strength has become excessively large compared with its initial value.
The search is stopped when the following condition is satisfied for at least one $i \in \{1, \dots, n\}$:
$\sigma^{(t)} s_i^{(t)} \sqrt{[\vec{C}^{(t)}]_{i,i}} > \epsilon_{\texttt{tfacupx}} \, \sigma^{(1)} s_i^{(1)}$.
%

\vspace{0.3em}
\noindent \textbf{$\bullet$} \texttt{tolupsigma}
%
Let $\sigma^{(t)} / \max_{i \in \{1, \dots, n\}} d_i^{(t)}$ be the magnitude of the effective step-size after factoring out the elongation due to the shape of the distribution.
The search is stopped when $\sigma^{(t)} / \max_{i \in \{1, \dots, n\}} d_i^{(t)} > \epsilon_{\texttt{tsigma}} \sigma^{(1)}$.
Thus, \texttt{tolupsigma} stops the search when the current step-size $\sigma^{(t)}$ is considerably larger than the initial one $\sigma^{(1)}$.


\vspace{0.3em}
\noindent \textbf{$\bullet$} \texttt{tolconditioncov}
This criterion stops the search when the condition number of the covariance matrix $\vec{C}$ is too large.
Let $\mathrm{cond}(\vec{C})$ denote the condition number of $\vec{C}$, i.e., $\mathrm{cond}(\vec{C}) = \max_{i \in \{1, \dots, n\}} d_i^2/\min_{i \in \{1, \dots, n\}} d_i^2$.
The search is stopped when $\mathrm{cond}(\vec{C}) > \epsilon_{\texttt{tcond}}$.

%% file: setup.tex
\section{Setup}
\label{sec:setup}

This section describes the experimental setup.
We used the noiseless BBOB function set (\texttt{bbob})~\cite{hansen2012fun}, which consists of 24 diverse functions.
The 24 \texttt{bbob} functions are grouped into the following five categories: separable functions ($f_1, ...,  f_5$), functions with low or moderate conditioning ($f_6, ..., f_9$), functions with high conditioning and unimodal ($f_{10}, ..., f_{14}$), multimodal functions with adequate global structure ($f_{15}, ..., f_{19}$), and multimodal functions with weak global structure ($f_{20}, ...,  f_{24}$).
The dimension $n$ was set to $2, 3, 5, 10, 20$ and $40$. 
According to the COCO platform~\cite{HansenARMTB21}, the number of instances was set to 15 for each function.
In other words, 15 independent runs were performed for each function.
The optimal solution of each \texttt{bbob} function instance lies in  $[-5,5]^n$.


All experiments were conducted using \texttt{pycma}~\cite{Hansen19} version \texttt{4.4.4}.
In CMA-ES, the default sample size is given by $\lambda_{\text{def}} = 4 + \lfloor 3 \ln(n) \rfloor$.
This work considers the following nine settings of $\lambda$: $1\lambda_{\text{def}}$, $2\lambda_{\text{def}}$, $4\lambda_{\text{def}}$, $8\lambda_{\text{def}}$, $16\lambda_{\text{def}}$, $32\lambda_{\text{def}}$, $64\lambda_{\text{def}}$, $128\lambda_{\text{def}}$, and $256\lambda_{\text{def}}$.
As in \cite{Hansen09}, the initial mean vector $\vec{m}^{(1)}$ was generated uniformly at random from $[-4, 4]^n$.
The initial step-size $\sigma^{(1)}$ was set to $2$, which is one-fifth of the width of the search space.
Let $\vec{x}^{\text{bsf}}$ and $\vec{x}^{*}$ denote the best-so-far solution found by CMA-ES and the optimal solution, respectively.
If $|f(\vec{x}^{\text{bsf}}) - f(\vec{x}^{*})| \leq 10^{-8}$, we replace $f(\vec{x}^{\text{bsf}})$ with $f(\vec{x}^{*})$.

All parameters of the 12 stopping criteria were set to their default values shown in \pref{tab:sc_parameters}.
The maximum number of function evaluations was set to $10^5 \times n$. 
\textit{Each run of CMA-ES continued until the maximum number of function evaluations was reached or the \texttt{pycma} implementation crashed, even if a stopping criterion had been triggered earlier}.
Even when CMA-ES reached the optimal solution, the search was not stopped.
This is because we wanted to examine whether the stopping criteria could detect convergence.
%
As described in the following section, the performance of each stopping criterion is discussed based on the number of function evaluations at which it was triggered.

\subsection{How to evaluate the performance of stopping criteria}
\label{sec:pose}


A previous study~\cite{KitamuraT26} proposed a \underline{p}erformance measure based on the \underline{o}ptimal number of function evaluations for \underline{s}topping \underline{e}volutionary algorithms (POSE).
The POSE measure was originally designed to quantitatively evaluate the performance of stopping criteria for evolutionary multi-objective optimization (EMO).
Fortunately, the POSE measure can be straightforwardly extended to evaluate the performance of stopping criteria for single-objective black-box optimization algorithms, including CMA-ES.
In the following, we describe this extended version of the POSE measure.

Let $\texttt{FE}^{\mathrm{stop}}$ be the number of function evaluations when a stopping criterion actually stops the search.
Let also $\texttt{FE}^{*}$ be the optimal number of function evaluations to stop the search in a single run.
The previous work~\cite{KitamuraT26} defines $\texttt{FE}^{*}$ as the number of function evaluations at which the best-so-far hypervolume~\cite{ZitzlerT98} value is last updated in a single run of an EMO algorithm.\footnote{The previous work~\cite{KitamuraT26} introduced a parameter $\delta$ that determines the update range of the hypervolume values between two consecutive iterations.
Since the default value of $\delta$ is zero, $\delta$ has no effect in our setting.
Therefore, we omit $\delta$.
}
In this work, $\texttt{FE}^{*}$ is simply defined as the number of function evaluations at which the best-so-far objective value was last updated in a single run of CMA-ES.

\begin{figure}[t]
\newcommand{\wir}{0.99}
\centering
\includegraphics[width=\wir\textwidth]{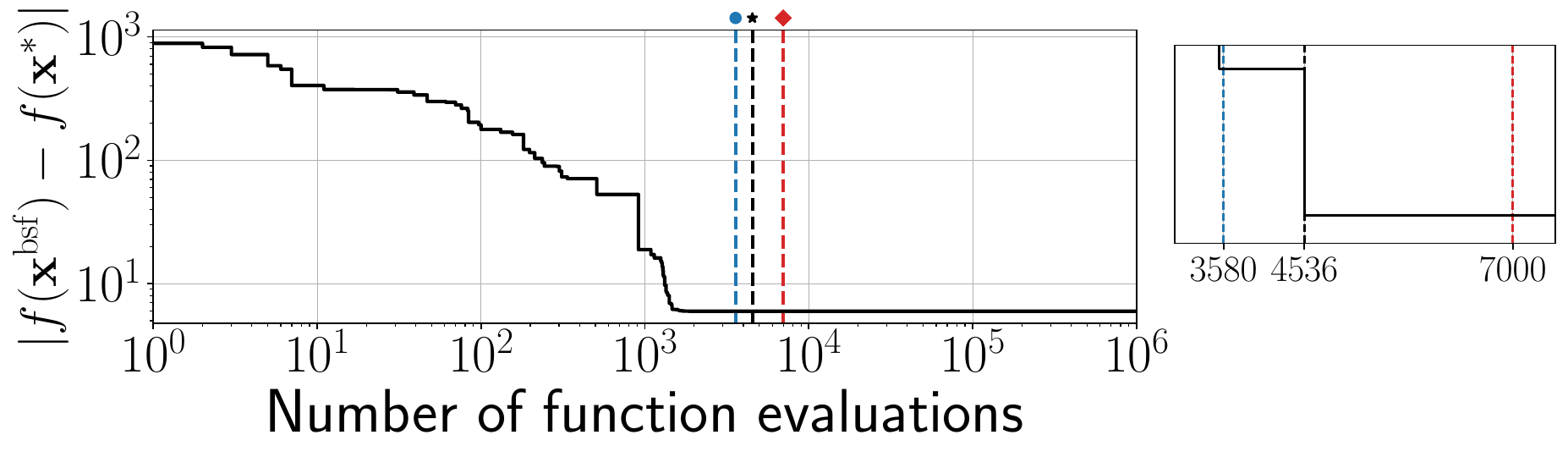}
\caption{$\texttt{FE}^{*}$ ($\bigstar$) and $\texttt{FE}^{\text{stop}}$ of 
\texttt{tolfun} (\tabblue{$\bullet$}) and \texttt{tolstagnation} (\tabred{$\blacklozenge$}) in a single run of CMA-ES. The black line shows the error value $|f(\mathbf{x}^{\mathrm{bsf}}) -  f(\mathbf{x}^*)|$.}
\label{fig:example_festar}
\end{figure}

\pref{fig:example_festar} shows an example of $\texttt{FE}^{*}$ and $\texttt{FE}^{\text{stop}}$ of two stopping criteria (\texttt{tolfun} and \texttt{tolstagnation}) in a single run of CMA-ES.
\pref{fig:example_festar} shows the results on an instance of $f_3$ with $n=10$.
In \pref{fig:example_festar}, the horizontal axis represents the number of function evaluations on a logarithmic scale, and the vertical axis represents the error value $|f(\vec{x}^{\text{bsf}}) - f(\vec{x}^{*})|$, where $\vec{x}^{\text{bsf}}$ is the best-so-far solution and $\vec{x}^{*}$ is the optimal solution.
As shown in \pref{fig:example_festar}, CMA-ES does not reach the optimal solution $\vec{x}^{*}$.

In \pref{fig:example_festar}, $\texttt{FE}^{*}$ is $4\,536$, and the $\texttt{FE}^{\text{stop}}$ values of \texttt{tolfun} and \texttt{tolstagnation} are $3\,580$ and $7\,000$, respectively.
In this run, \texttt{tolfun} is triggered early, whereas \texttt{tolflatfitness} is triggered late.
Since the \texttt{tolfun} criterion stops the search before $\texttt{FE}^{*}$, the quality of the best-so-far solution $\vec{x}^{\text{bsf}}$ is lower than that at $\texttt{FE}^{*}$.
In other words, the \texttt{tolfun} criterion stops the search before CMA-ES reaches complete stagnation.
In contrast to \texttt{tolfun}, \texttt{tolstagnation} stops the search after $\texttt{FE}^{*}$.
In this case, $\texttt{FE}^{\text{stop}} - \texttt{FE}^{*}$ function evaluations are wasted because no further improvement in the best-so-far solution $\vec{x}^{\text{bsf}}$ occurs after $\texttt{FE}^{*}$.
Note that $\texttt{FE}^{*}$ is defined separately for each run in an a posteriori manner.
Thus, $\texttt{FE}^{*}$ depends strongly on the behavior of CMA-ES.

The POSE measure quantitatively evaluates the performance of a stopping criterion based on the difference between $\texttt{FE}^{\mathrm{stop}}$ and $\texttt{FE}^{*}$ as follows:
\begin{align}
\label{eq:pose}
\text{POSE} =
\begin{cases}
  \frac{|\texttt{FE}^{*} - \texttt{FE}^{\mathrm{stop}}|}{\texttt{FE}^{\mathrm{max}}} & \text{if } \texttt{FE}^{\mathrm{stop}} \geq \texttt{FE}^{*} \\
  \alpha\frac{|\texttt{FE}^{*} - \texttt{FE}^{\mathrm{stop}}|}{\texttt{FE}^{\mathrm{max}}} & \text{otherwise}, 
\end{cases}
\end{align}
where $\texttt{FE}^{\mathrm{max}}$ is the maximum number of function evaluations.
The difference $|\texttt{FE}^{*} - \texttt{FE}^{\mathrm{stop}}|$ is normalized by $\texttt{FE}^{\mathrm{max}}$.
A penalty factor $\alpha \geq 1 $ penalizes a criterion that stops before $\texttt{FE}^{*}$.
In this work, we set $\alpha = 1$, meaning that no such penalty is applied.
%
A small POSE value indicates that the corresponding stopping criterion stops the search at an appropriate point.
When a stopping criterion is triggered exactly at $\texttt{FE}^{*}$, its POSE value is $0$.

\newcommand{\colortextbox}[2]{%
  \begingroup
  \setlength{\fboxsep}{1pt}%
  \colorbox{#1}{\strut #2}%
  \endgroup
}

%% file: results.tex
\section{Results}
\label{sec:results}

This section describes our results.
Through experiments, Sections \ref{sec:ftc}--\ref{sec:results_early} aim to address the following three research questions, respectively:

\begin{enumerate}[RQ1:]
\item Which stopping criterion is most frequently triggered first in CMA-ES? 
\item How accurately does each stopping criterion stop the search? 
\item How frequently do the stopping criteria stop the search before $\texttt{FE}^*$? 
\end{enumerate}


Note that this paper focuses only on the effectiveness of stopping criteria for CMA-ES rather than on the performance of CMA-ES itself.
Note also that, as a first step, we conducted experiments using the default settings of the stopping criteria shown in \pref{tab:sc_parameters}.
Future work should investigate how hyperparameter settings affect the performance of stopping criteria for CMA-ES.



\subsection{Stopping criterion most frequently triggered first}
\label{sec:ftc}

\begin{figure}[t]
\newcommand{\wir}{0.33}
\centering
\includegraphics[width=0.99\textwidth]{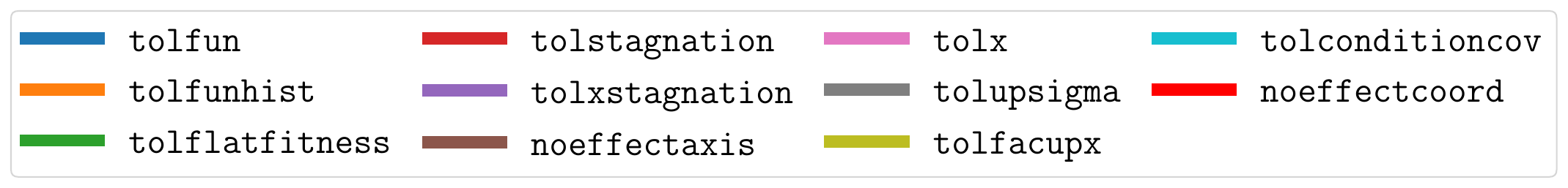}
\\
\vspace{-0.9em}
\subfloat[$\lambda=1 \lambda_{\text{def}}$]{\includegraphics[width=\wir\textwidth]{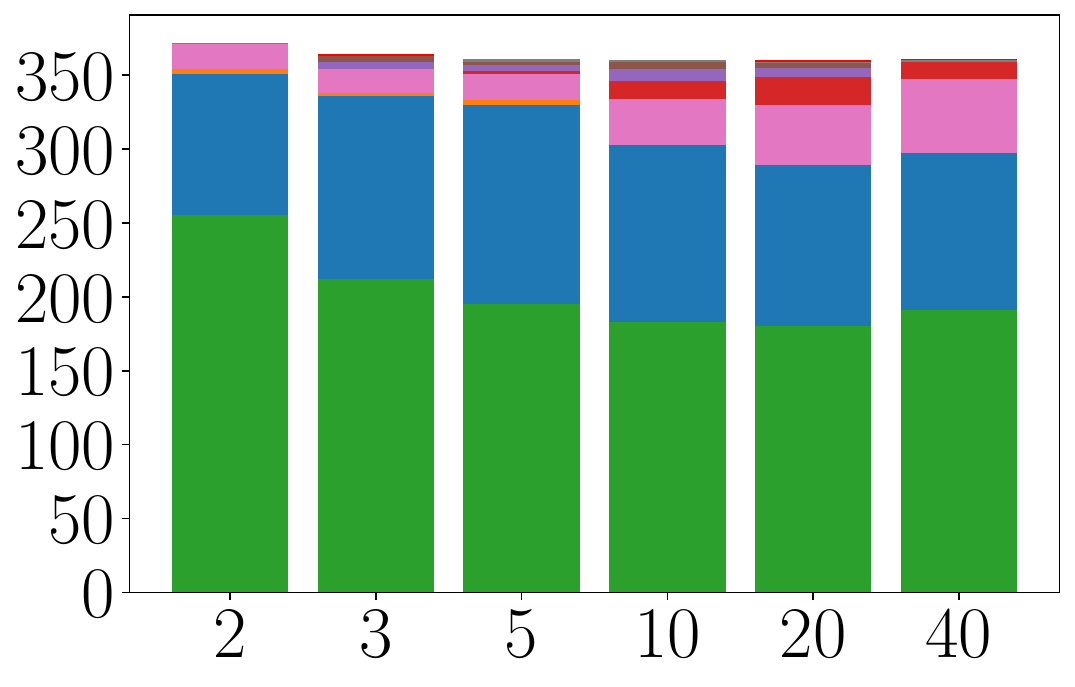}}
\subfloat[$\lambda=2 \lambda_{\text{def}}$]{\includegraphics[width=\wir\textwidth]{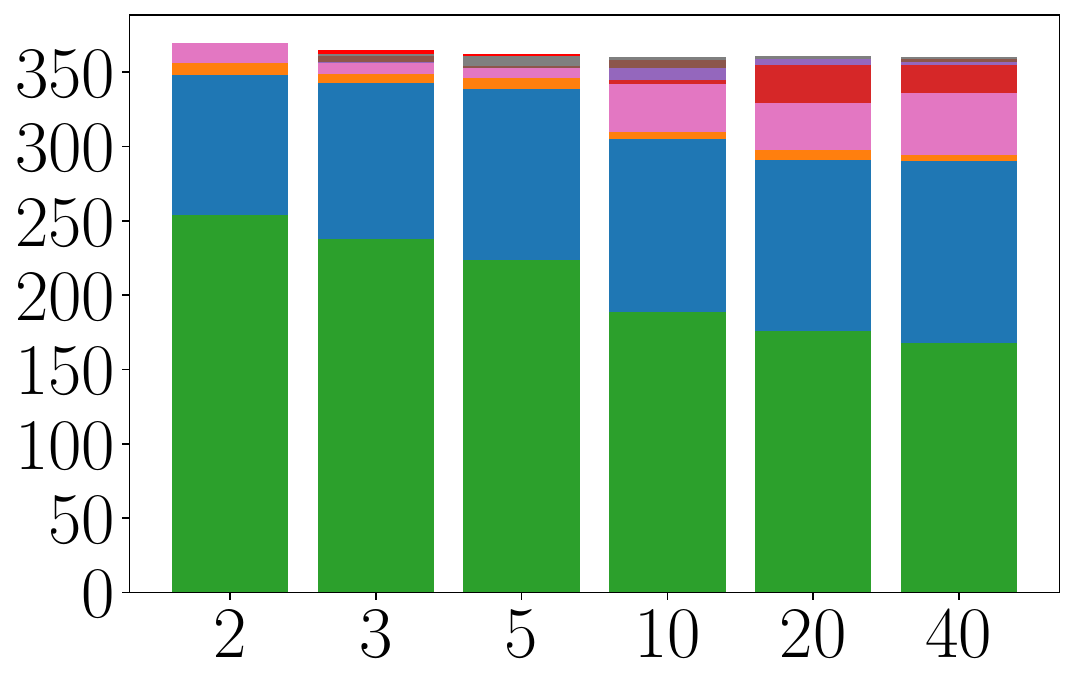}}
\subfloat[$\lambda=4 \lambda_{\text{def}}$]{\includegraphics[width=\wir\textwidth]{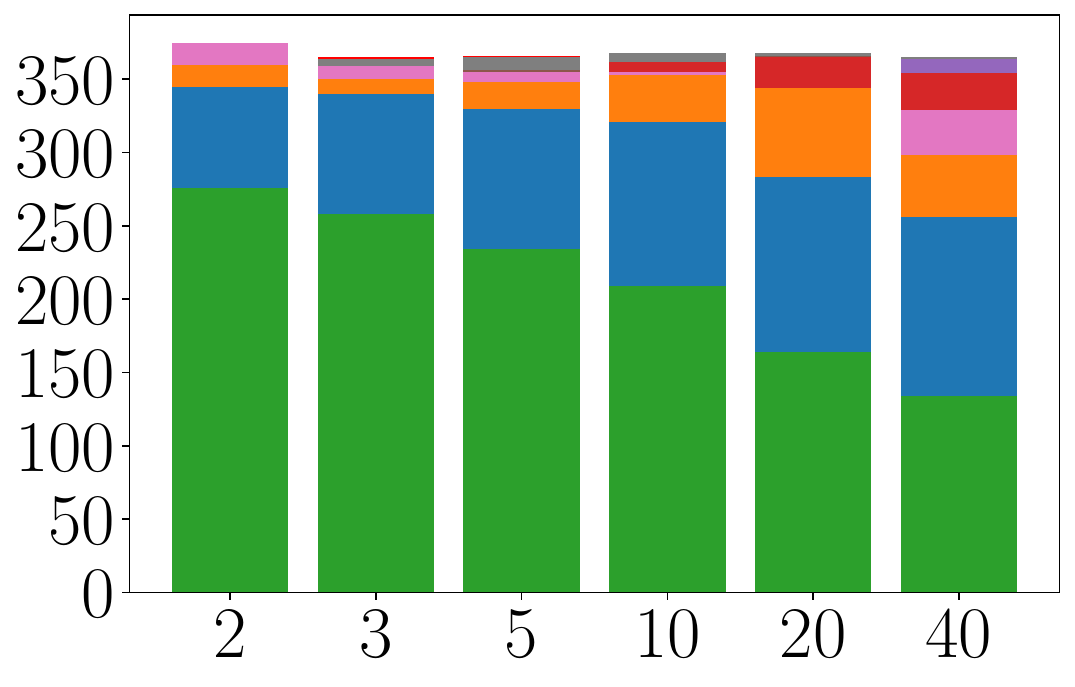}}
\\[-0.6em]
\subfloat[$\lambda=8  \lambda_{\text{def}}$]{\includegraphics[width=\wir\textwidth]{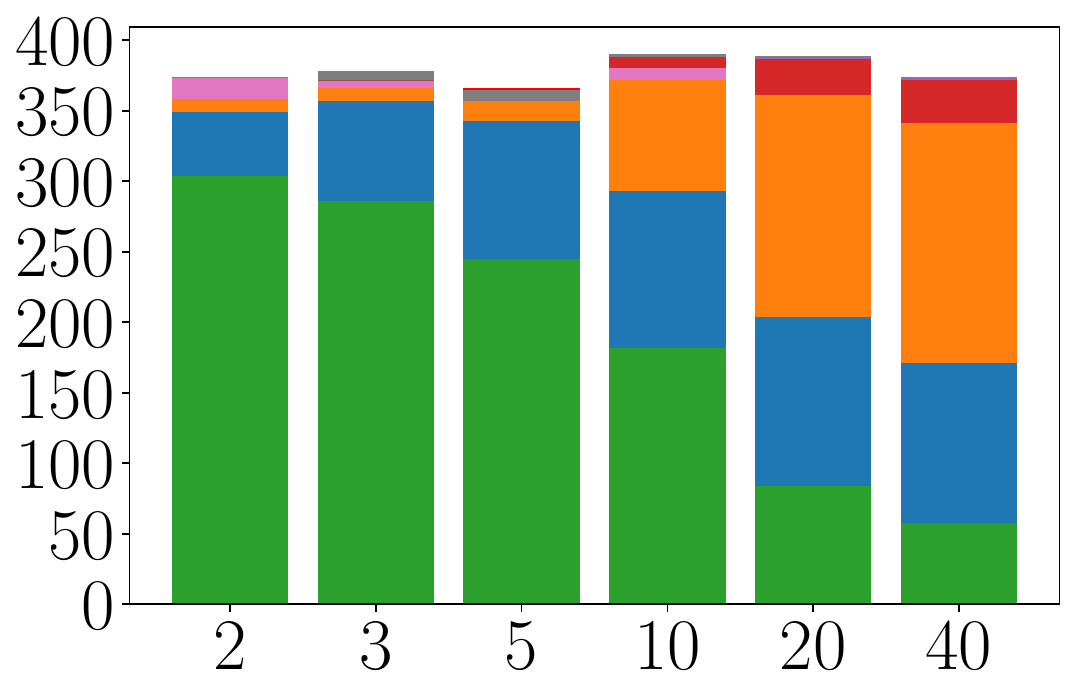}}
\subfloat[$\lambda=16  \lambda_{\text{def}}$]{\includegraphics[width=\wir\textwidth]{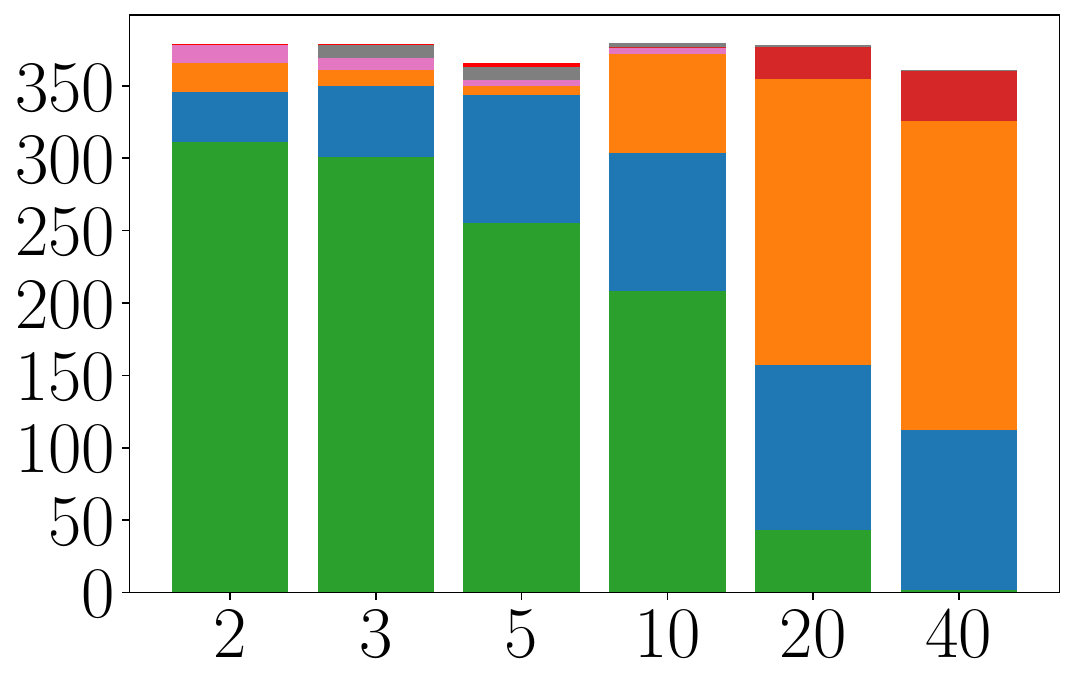}}
\subfloat[$\lambda=32  \lambda_{\text{def}}$]{\includegraphics[width=\wir\textwidth]{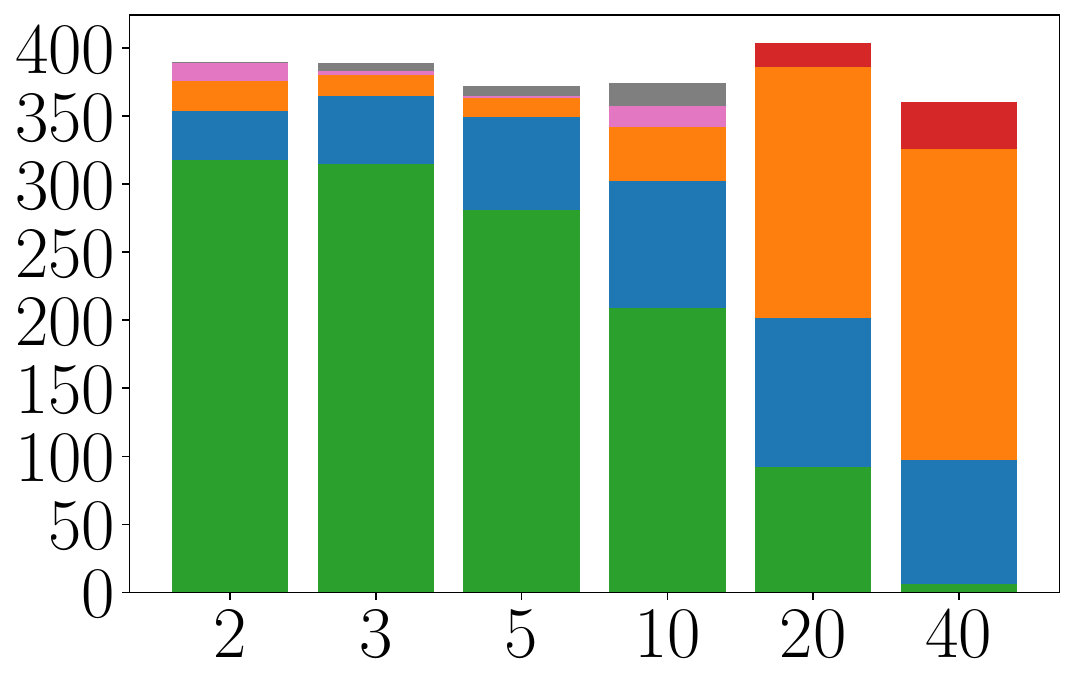}}
\\[-0.6em]
\subfloat[$\lambda=64  \lambda_{\text{def}}$]
{\includegraphics[width=\wir\textwidth]{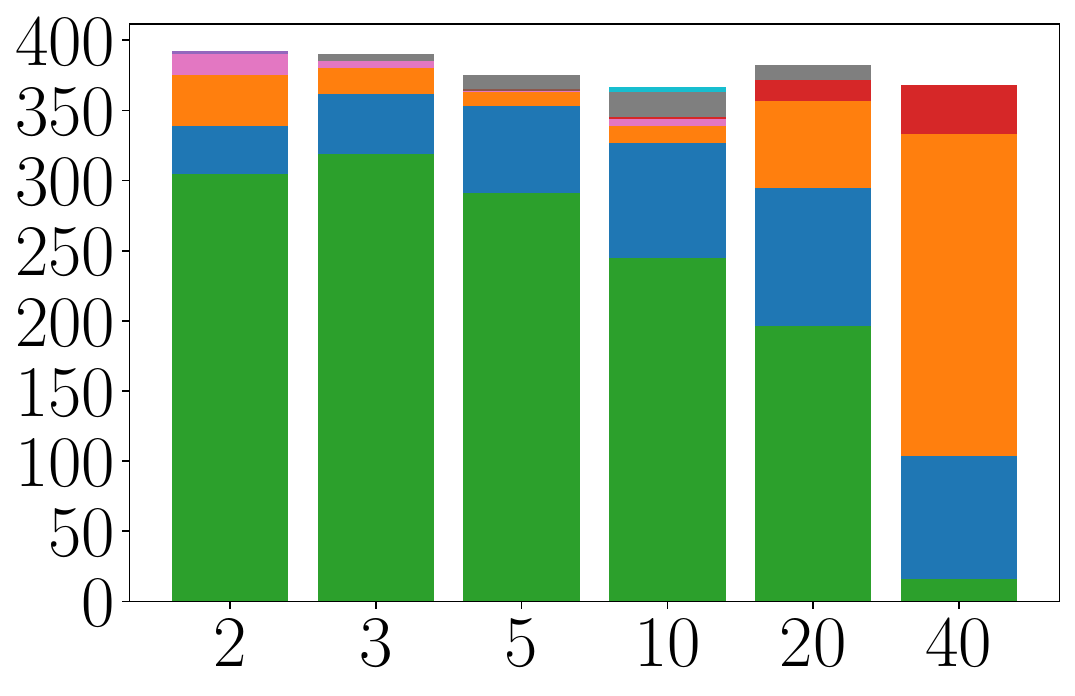}}
\subfloat[$\lambda=128  \lambda_{\text{def}}$]
{\includegraphics[width=\wir\textwidth]{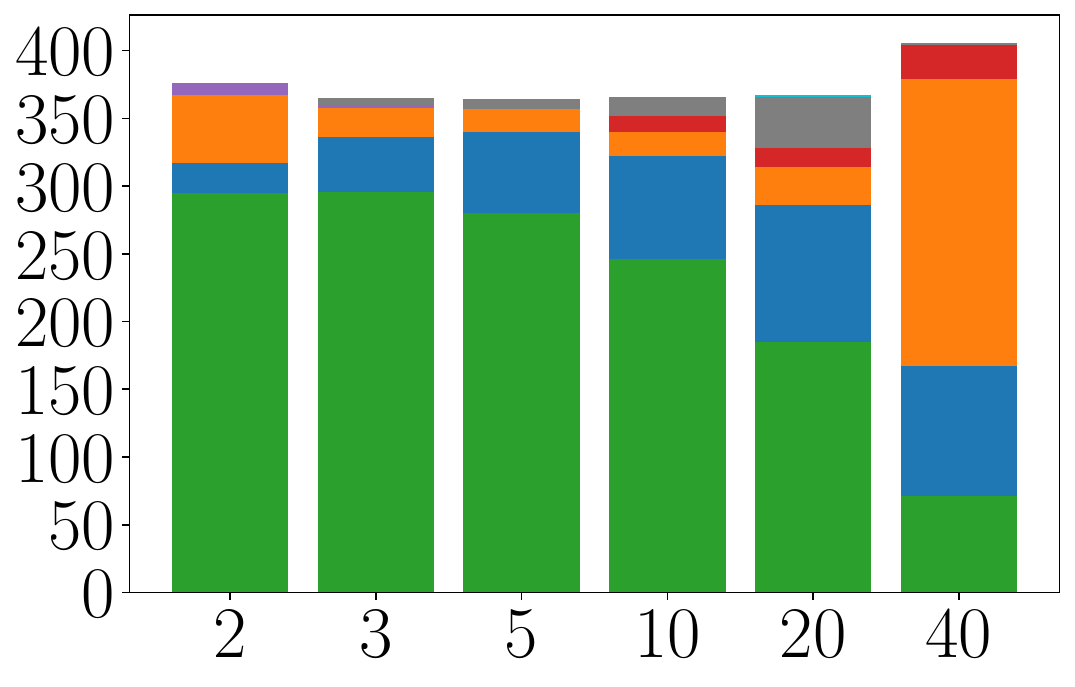}}
\subfloat[$\lambda=256  \lambda_{\text{def}}$]
{\includegraphics[width=\wir\textwidth]{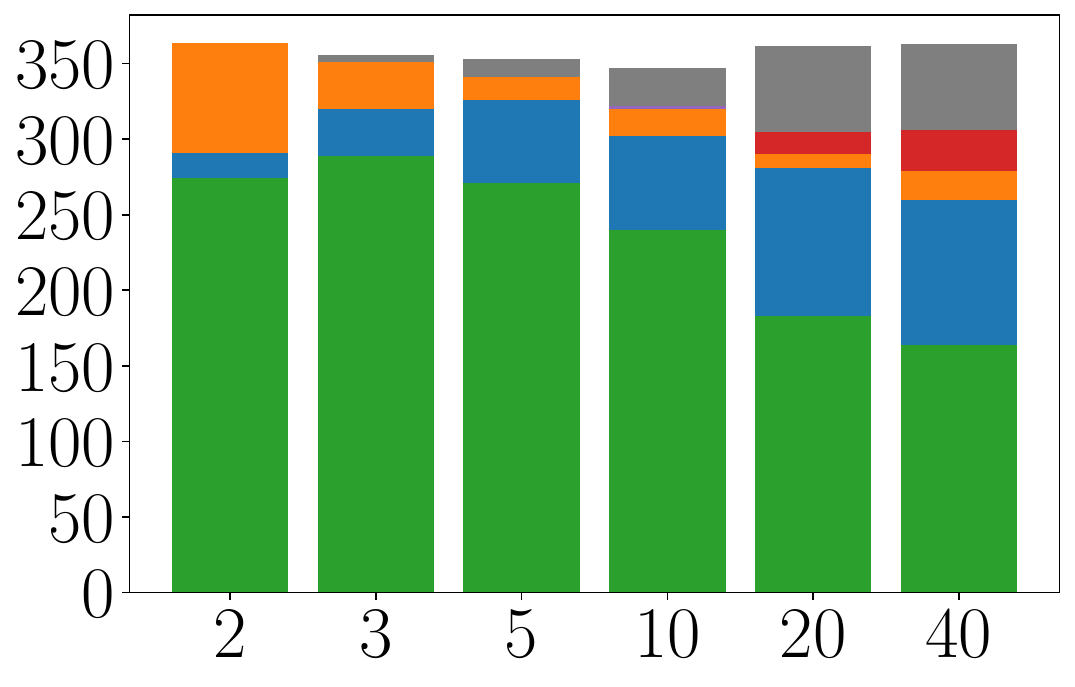}}
\caption{Number of times each stopping criterion was triggered first among the 11 criteria. The \textbf{horizontal axis} shows the dimension, $n \in \{2, 3, 5, 10, 20, 40\}$, and the \textbf{vertical axis} shows the frequency.}
\label{fig:ftc}
\end{figure}

\pref{fig:ftc} shows the number of times each stopping criterion was the first to be triggered among the 11 stopping criteria.
\pref{fig:ftc} shows the results for the nine $\lambda$ settings over the 360 $(=24 \times 15)$ function instances for each dimension $n$.
\pref{fig:ftc} shows which stopping criterion stops the CMA-ES search most frequently.
For example, as shown in \pref{fig:ftc}(a), the \texttt{tolflatfitness} criterion stops the search in about $250$ of the $360$ runs.
Note that more than one stopping criterion can be triggered simultaneously.
In addition, we observed that no stopping criterion stopped the search on some instances of $f_{13}$, $f_{23}$, and $f_{24}$ when $\lambda$ was large.
For these reasons, the total count in each stacked bar is not always 360.
Since the \texttt{noeffectaxis} and \texttt{tolfacupx} criteria were never the first to be triggered in our experiments, they are not shown in \pref{fig:ftc}.
The \texttt{tolxstagnation} and \texttt{tolconditioncov} criteria were also almost never the first to be triggered.

As shown in \pref{fig:ftc}, the sample size $\lambda$ and the dimension $n$ influence how frequently each stopping criterion is triggered.
\pref{fig:ftc} suggests that \texttt{tolflatfitness} stops the search most frequently in many cases, followed by \texttt{tolfun}. 
The \texttt{tolx} criterion is occasionally the earliest stopping criterion to become active for $\lambda \in \{1\lambda_{\text{def}}, 2\lambda_{\text{def}}\}$, especially in high dimensions.
In other words, \texttt{tolx} is almost never triggered before the other criteria for larger values of $\lambda$.
As seen from Figs. \ref{fig:ftc}(c)--(h), \texttt{tolfunhist} is frequently the first criterion to be triggered as $\lambda$ and $n$ increase.
In particular, as shown in Figs. \ref{fig:ftc}(d)--(h), \texttt{tolfunhist}  most frequently stops the search for $n=40$ when $\lambda \in \{8\lambda_{\text{def}}, 16\lambda_{\text{def}}, 32\lambda_{\text{def}}, 64\lambda_{\text{def}} \}$.
Similar to \texttt{tolfunhist}, as shown in Figs. \ref{fig:ftc}(h) and (i), \texttt{tolupsigma} is more likely to be activated first as $n$ and $\lambda$ increase.





\subsubsection*{Summary and discussion}

Our results show that the stopping criterion triggered first depends on $\lambda$ and $n$. While some fitness-based criteria (e.g., \texttt{tolflatfitness} and \texttt{tolfun}) are frequently triggered first, convergence-based and divergence-based criteria rarely are, except for \texttt{tolupsigma}.


Although CMA-ES uses a portfolio of multiple stopping criteria, it has been unclear which criterion is most frequently activated within that portfolio. 
Our results help to address this overlooked issue.
Our results also indirectly contribute to understanding the behavior of IPOP-CMA-ES, which doubles the sample size $\lambda$ at each restart.
For example, as seen from \pref{fig:ftc}(a), the CMA-ES search with the initial $\lambda$ (i.e., $1\lambda_{\text{def}}$) is likely to be stopped by \texttt{tolflatfitness}.
As CMA-ES continues to restart, the search is likely to be stopped by \texttt{tolfunhist} for $n=40$.
Thus, different stopping criteria are likely to be triggered during a single run of IPOP-CMA-ES.

\begin{figure}[t]
\newcommand{\wir}{0.325}
\centering
\includegraphics[width=0.99\textwidth]{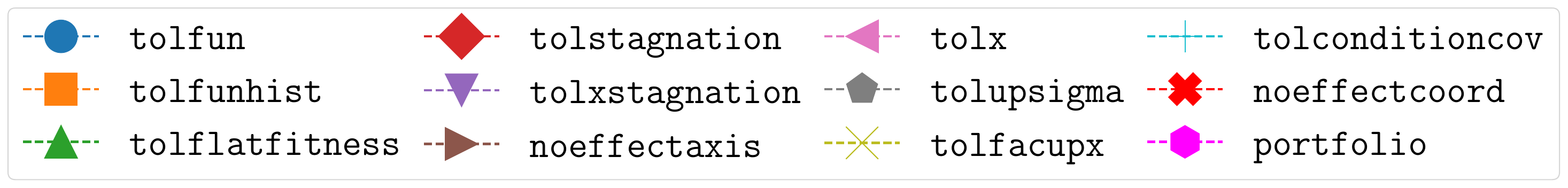}
\\
\vspace{-0.9em}
\subfloat[$\lambda=1 \lambda_{\text{def}}$]{\includegraphics[width=\wir\textwidth]{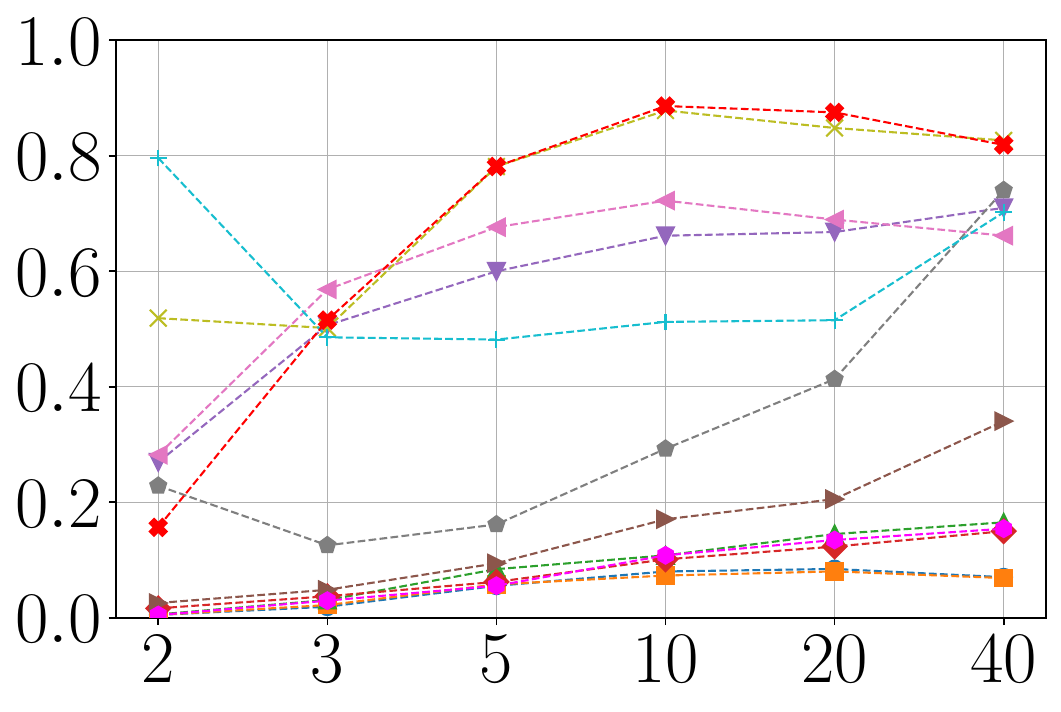}}
\subfloat[$\lambda=2 \lambda_{\text{def}}$]{\includegraphics[width=\wir\textwidth]{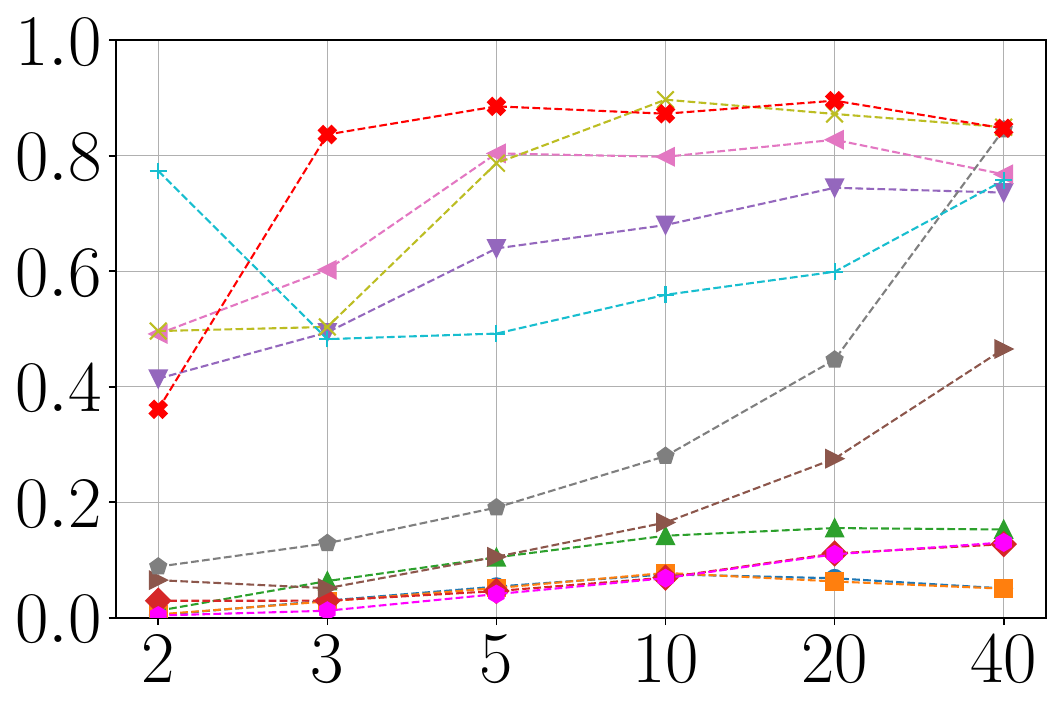}}
\subfloat[$\lambda=4 \lambda_{\text{def}}$]{\includegraphics[width=\wir\textwidth]{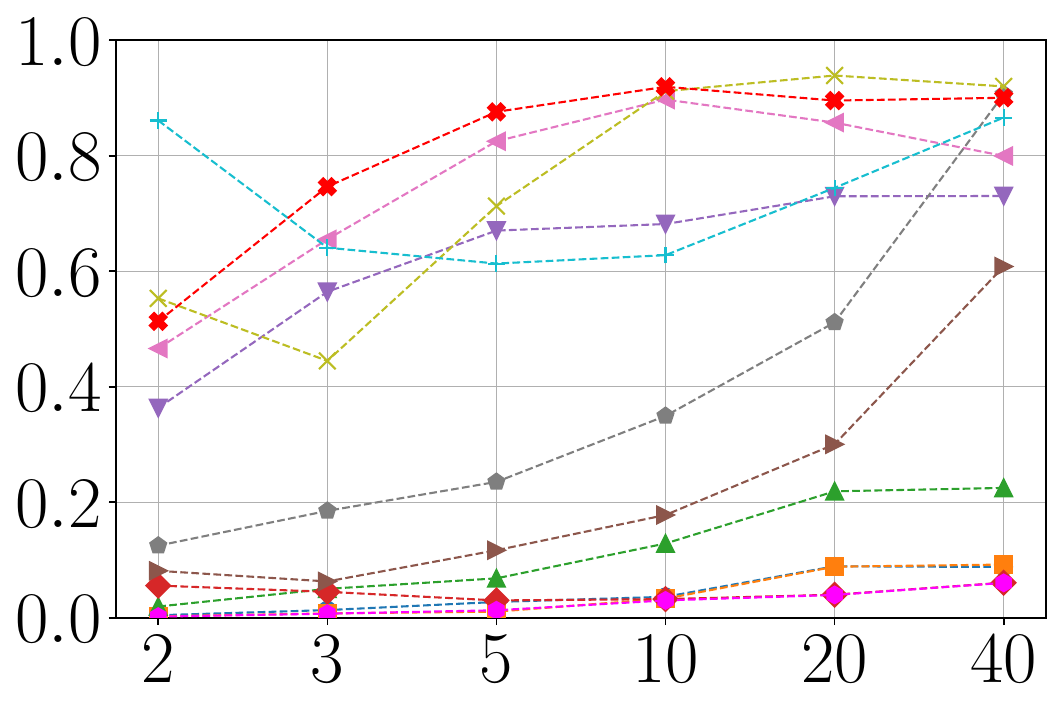}}
\\
\subfloat[$\lambda=8 \lambda_{\text{def}}$]{\includegraphics[width=\wir\textwidth]{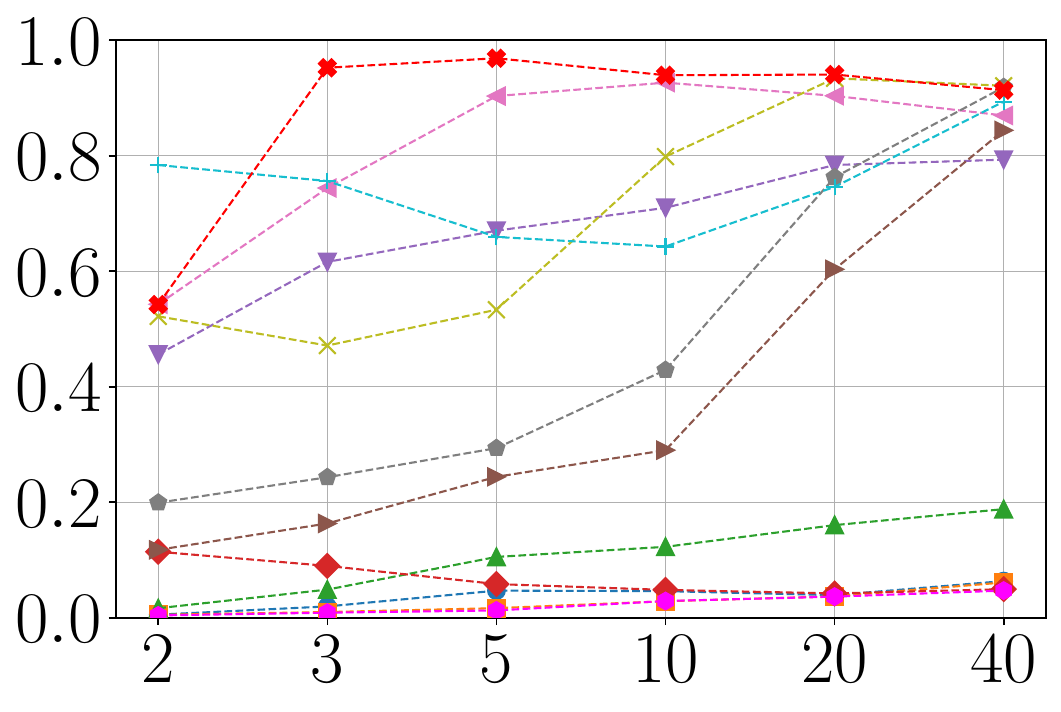}}
\subfloat[$\lambda=16 \lambda_{\text{def}}$]{\includegraphics[width=\wir\textwidth]{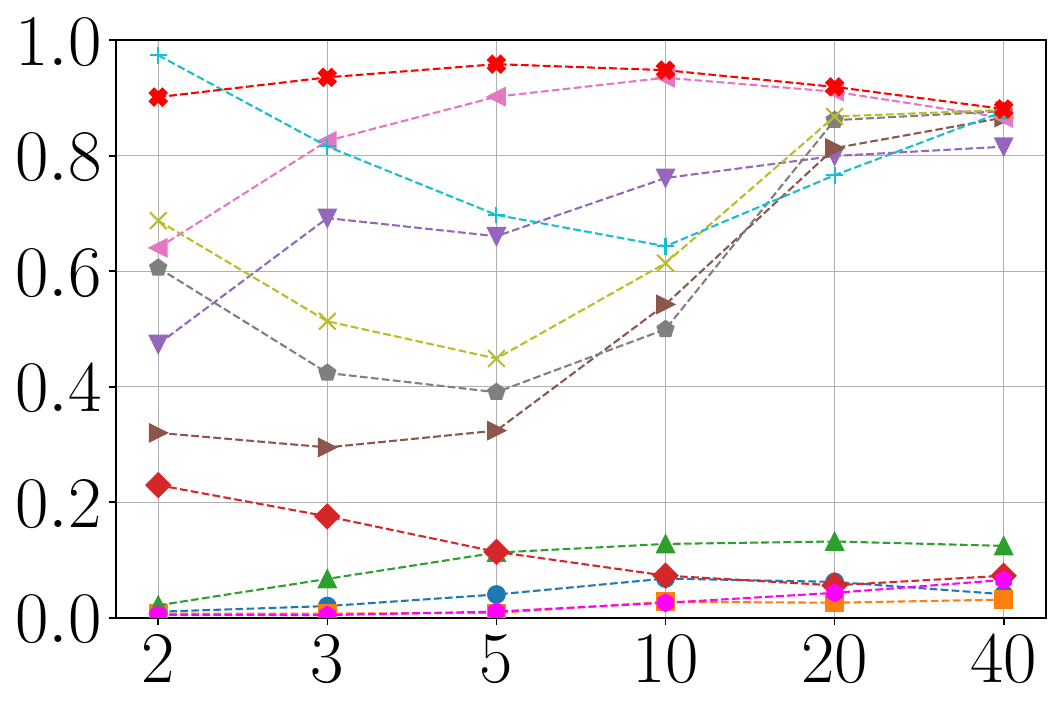}}
\subfloat[$\lambda=32 \lambda_{\text{def}}$]{\includegraphics[width=\wir\textwidth]{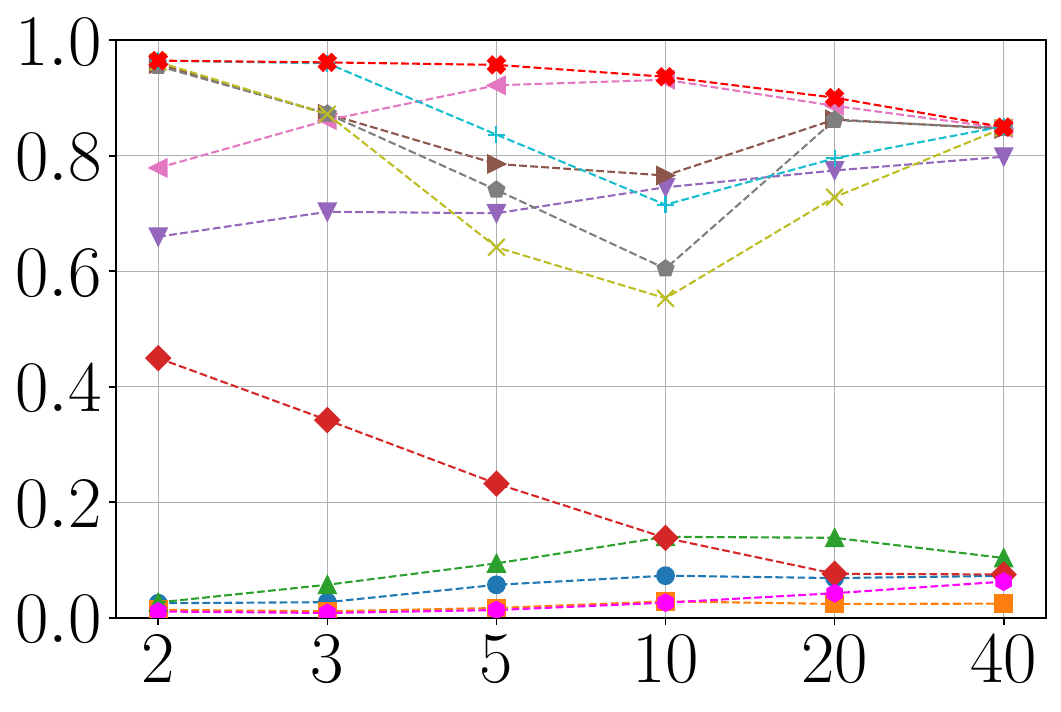}}
\\
\subfloat[$\lambda=64 \lambda_{\text{def}}$]{\includegraphics[width=\wir\textwidth]{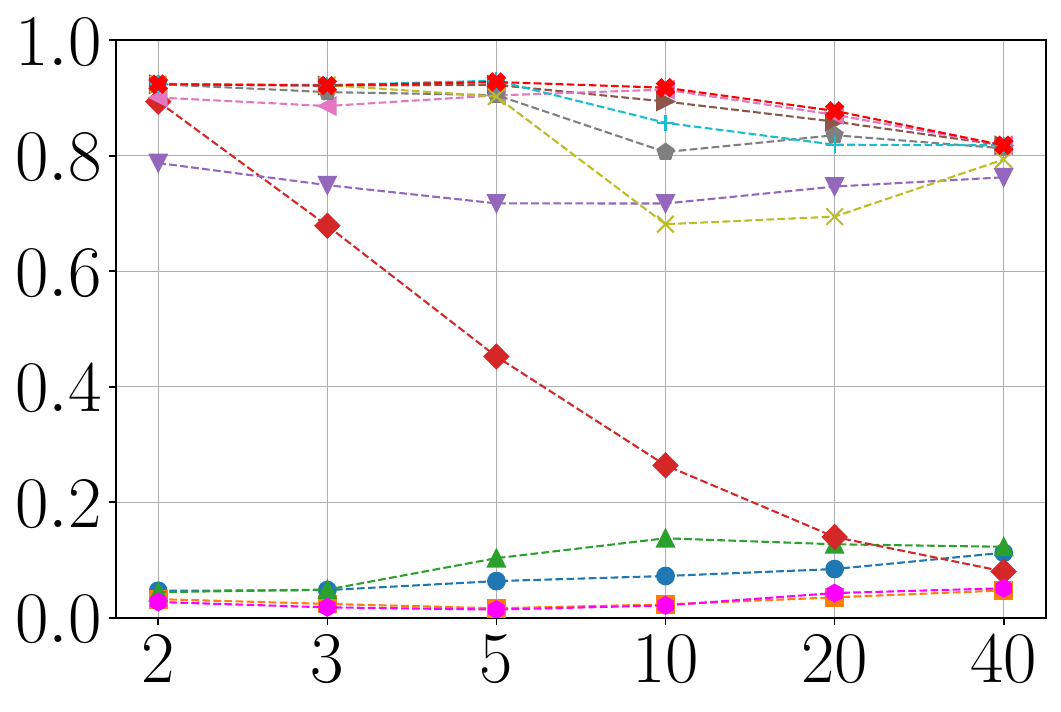}}
\subfloat[$\lambda=128 \lambda_{\text{def}}$]{\includegraphics[width=\wir\textwidth]{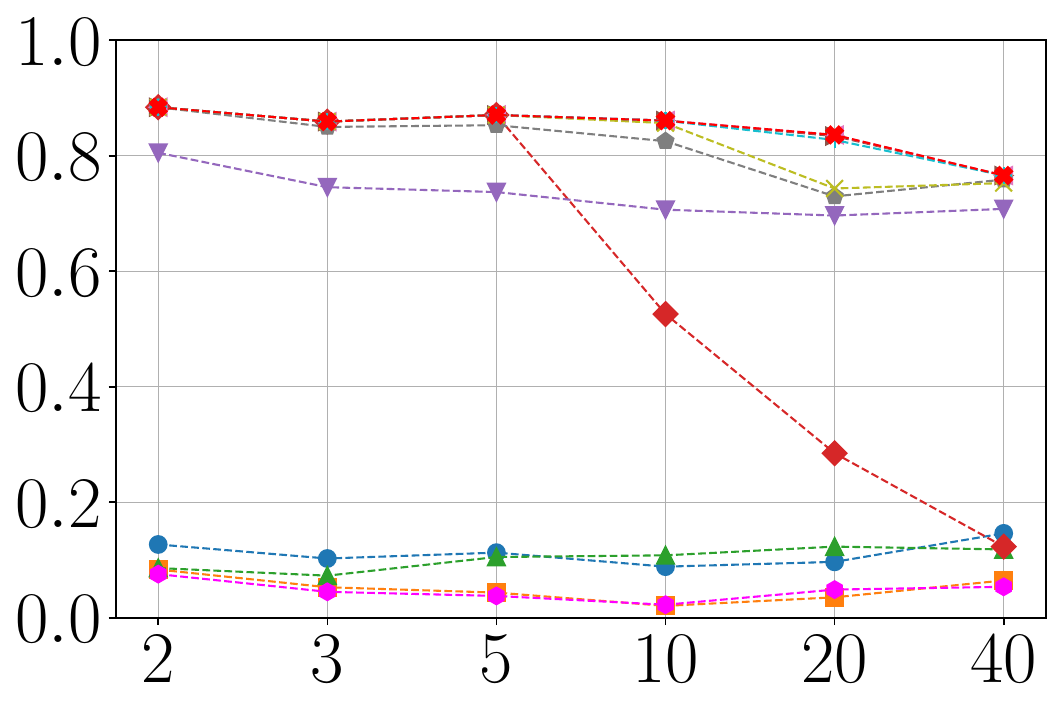}}
\subfloat[$\lambda=256 \lambda_{\text{def}}$]{\includegraphics[width=\wir\textwidth]{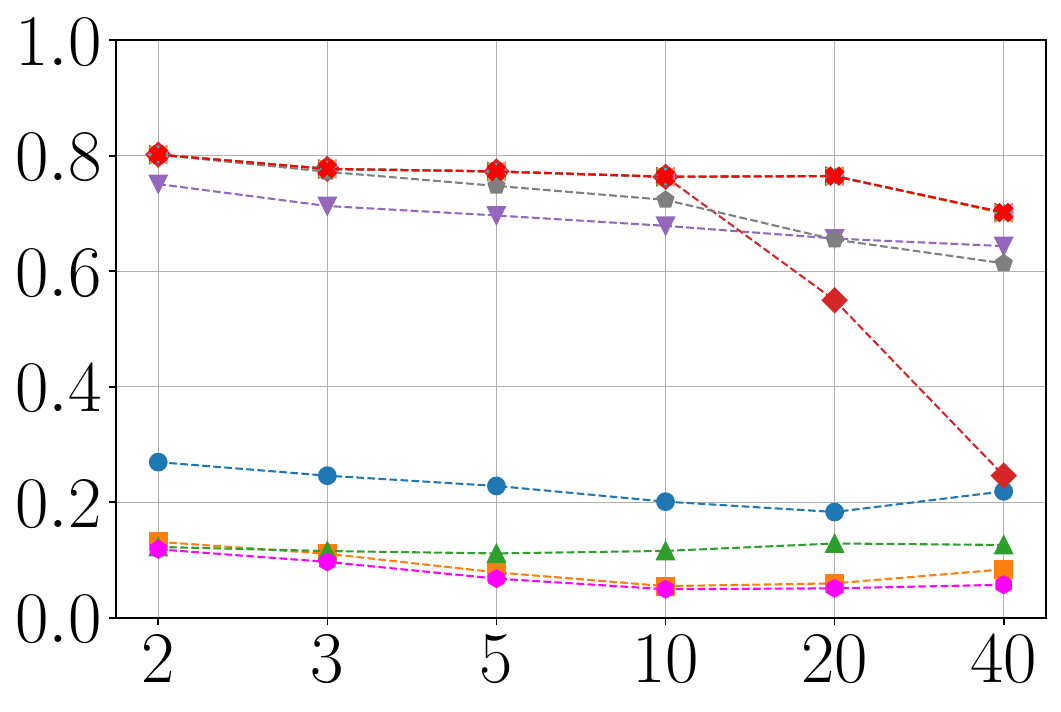}}
\caption{Average POSE values over the $360$ function instances. The \textbf{horizontal axis} shows the dimension, $n \in \{2, 3, 5, 10, 20, 40\}$, and the \textbf{vertical axis} shows the average POSE value.}
\label{fig:pose}
\end{figure}


\subsection{Accuracy of each criterion for stopping the CMA-ES search}
\label{sec:results_pose}

\pref{fig:pose} shows the average POSE values of the 11 stopping criteria and their portfolio, which stops the search when at least one criterion is triggered.
For details on which criterion actually stops the CMA-ES search within the portfolio, refer to \pref{sec:ftc}.
\pref{fig:pose} shows the results for the nine $\lambda$ settings over the 360 $(=24 \times 15)$ function instances for each dimension $n$.
When a stopping criterion is not activated in a single run, we set its $\texttt{FE}^{\text{stop}}$ to $\texttt{FE}^{\text{max}}$ in \pref{eq:pose}.
If CMA-ES crashes, $\texttt{FE}^{\text{stop}}$ is set to the number of function evaluations at the time of the crash.

\pref{fig:pose} shows how accurately each stopping criterion stops the CMA-ES search.
For example, as shown in \pref{fig:pose}(a), \texttt{tolconditioncov} achieves the worst performance for $\lambda=1 \lambda_{\text{def}}$ and $n=2$.
In contrast, \texttt{tolfun}, \texttt{tolfunhist}, and \texttt{portfolio} perform the best in this case.
As shown in \pref{fig:pose}, \texttt{tolfunhist} and \texttt{portfolio} obtain the best performance in most cases.
The performance of \texttt{noeffectaxis} is strongly affected by $\lambda$ and $n$.
The \texttt{noeffectaxis} criterion performs well for small values of $\lambda$ and $n$.
However, its performance deteriorates as $\lambda$ and $n$ increase.
The results for \texttt{tolstagnation} exhibit interesting behavior.
On the one hand, as shown in Figs. \ref{fig:pose}(a)--(c), the \texttt{tolstagnation} criterion performs well for $\lambda \in \{1 \lambda_{\text{def}}, 2 \lambda_{\text{def}}, 4 \lambda_{\text{def}}\}$.
On the other hand, as shown in Figs. \ref{fig:pose}(d)--(i), the performance of \texttt{tolstagnation} deteriorates in low-dimensional settings as $\lambda$ increases.
In contrast, the stopping accuracy of \texttt{tolstagnation} improves as $n$ increases.
%

\begin{figure}[t]
\newcommand{\wir}{0.325}
\centering
\includegraphics[width=0.99\textwidth]{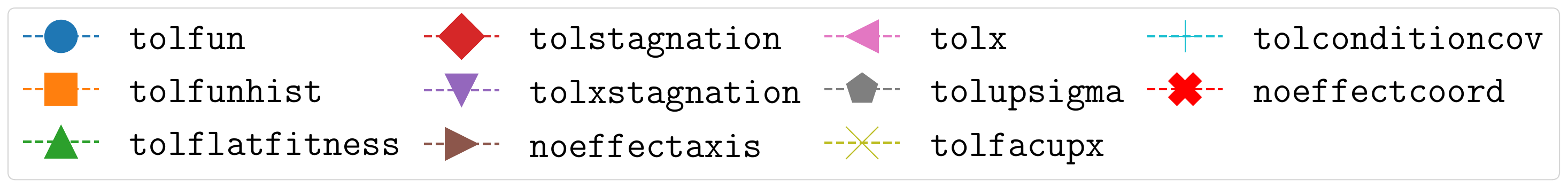}
\\
\vspace{-0.9em}
\subfloat[$\lambda=1 \lambda_{\text{def}}$]{\includegraphics[width=\wir\textwidth]{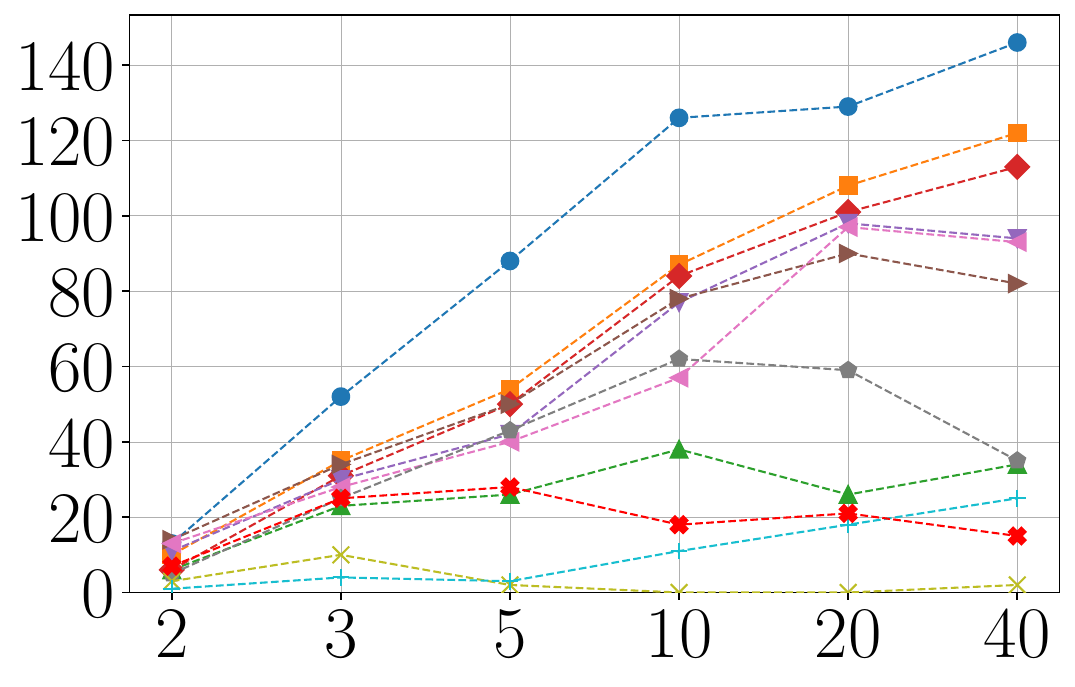}}
\subfloat[$\lambda=2 \lambda_{\text{def}}$]{\includegraphics[width=\wir\textwidth]{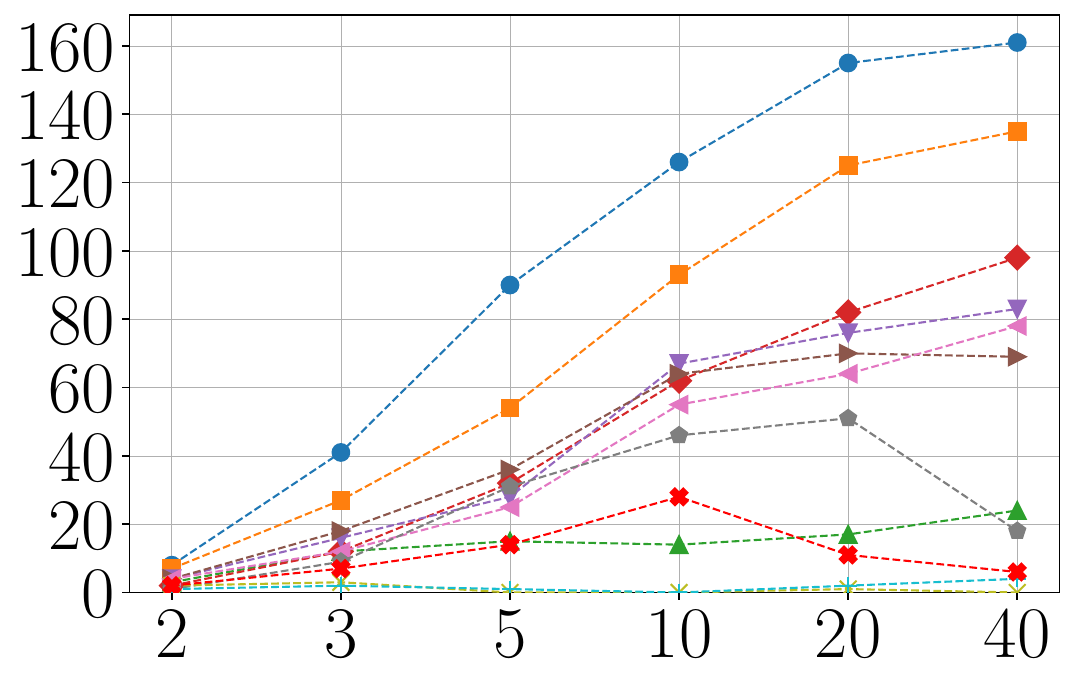}}
\subfloat[$\lambda=4 \lambda_{\text{def}}$]{\includegraphics[width=\wir\textwidth]{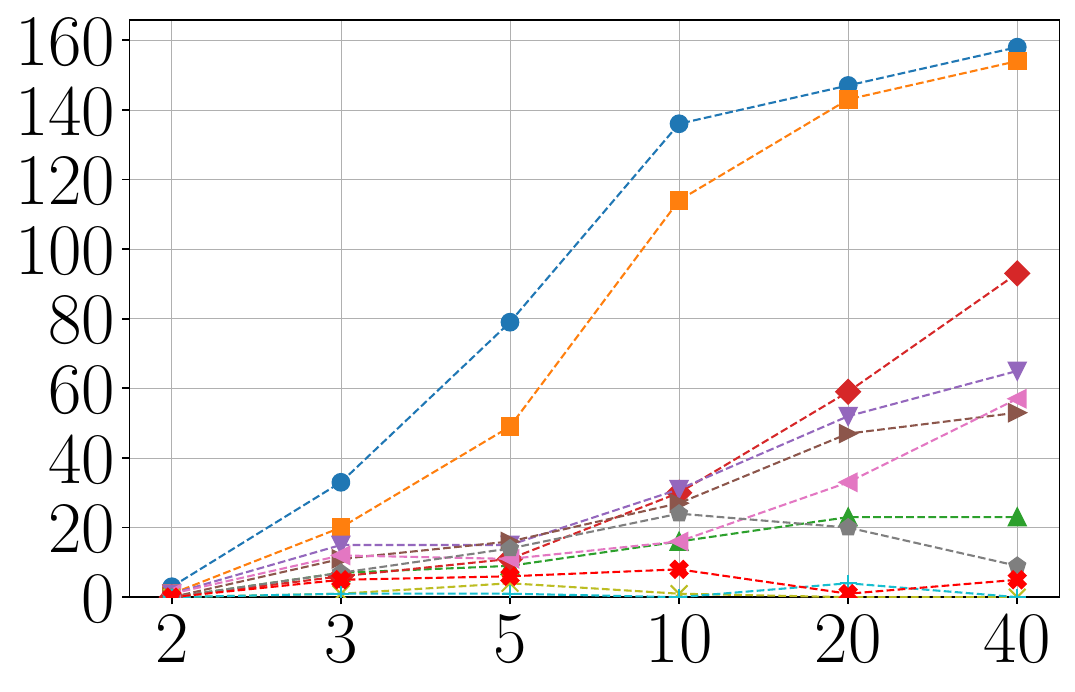}}
\\
\subfloat[$\lambda=8 \lambda_{\text{def}}$]{\includegraphics[width=\wir\textwidth]{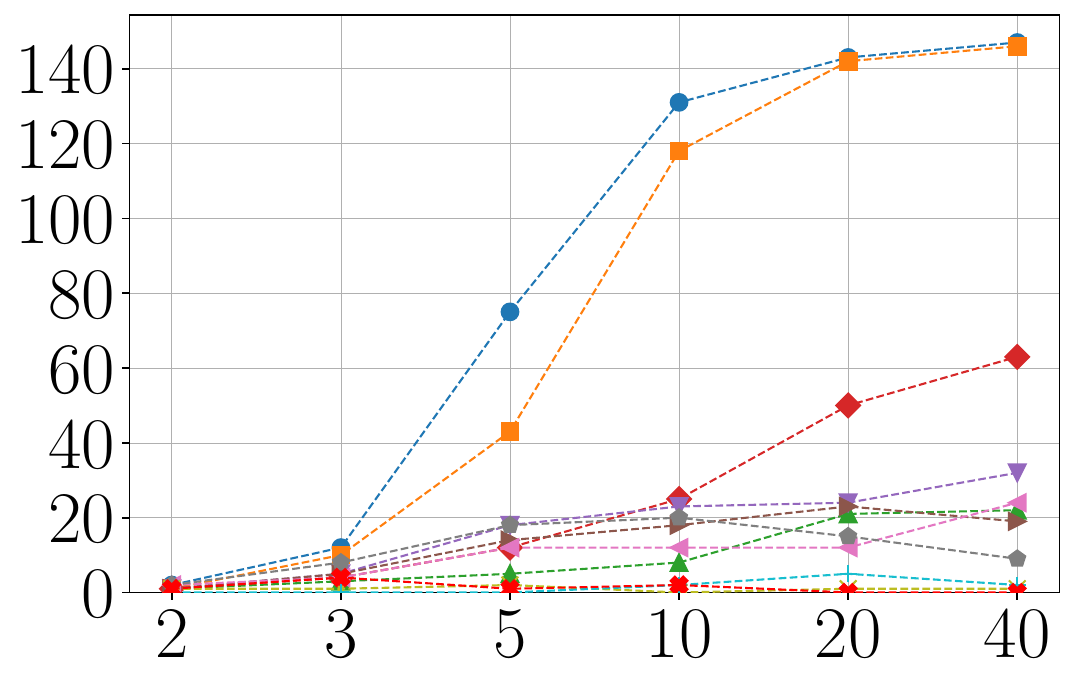}}
\subfloat[$\lambda=16 \lambda_{\text{def}}$]{\includegraphics[width=\wir\textwidth]{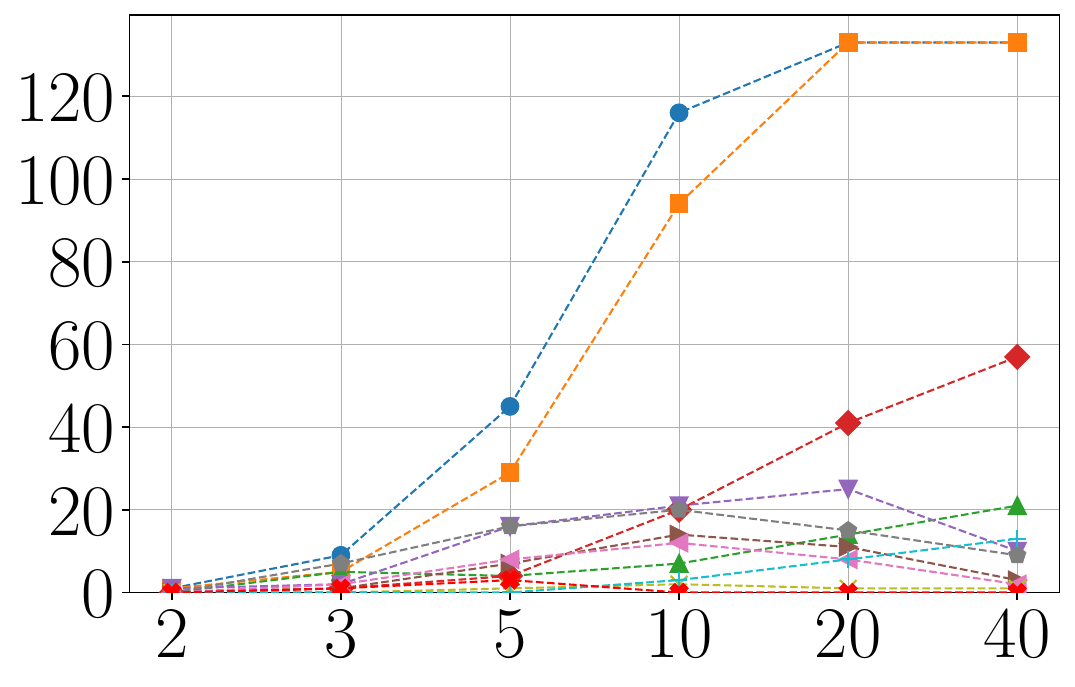}}
\subfloat[$\lambda=32 \lambda_{\text{def}}$]{\includegraphics[width=\wir\textwidth]{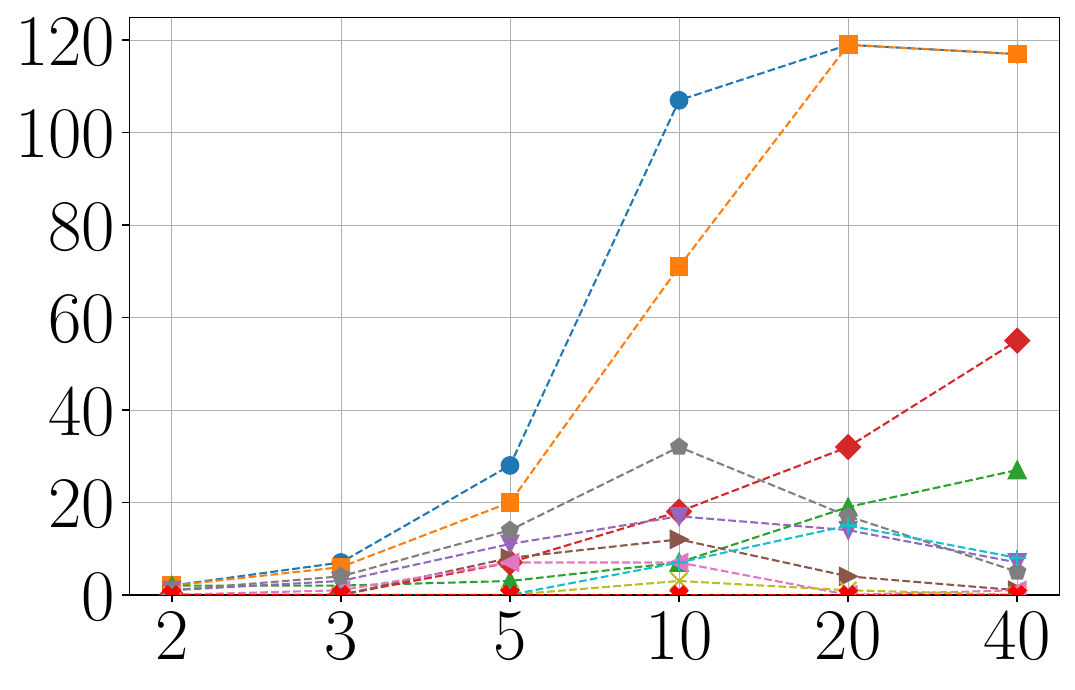}}
\\
\subfloat[$\lambda=64 \lambda_{\text{def}}$]{\includegraphics[width=\wir\textwidth]{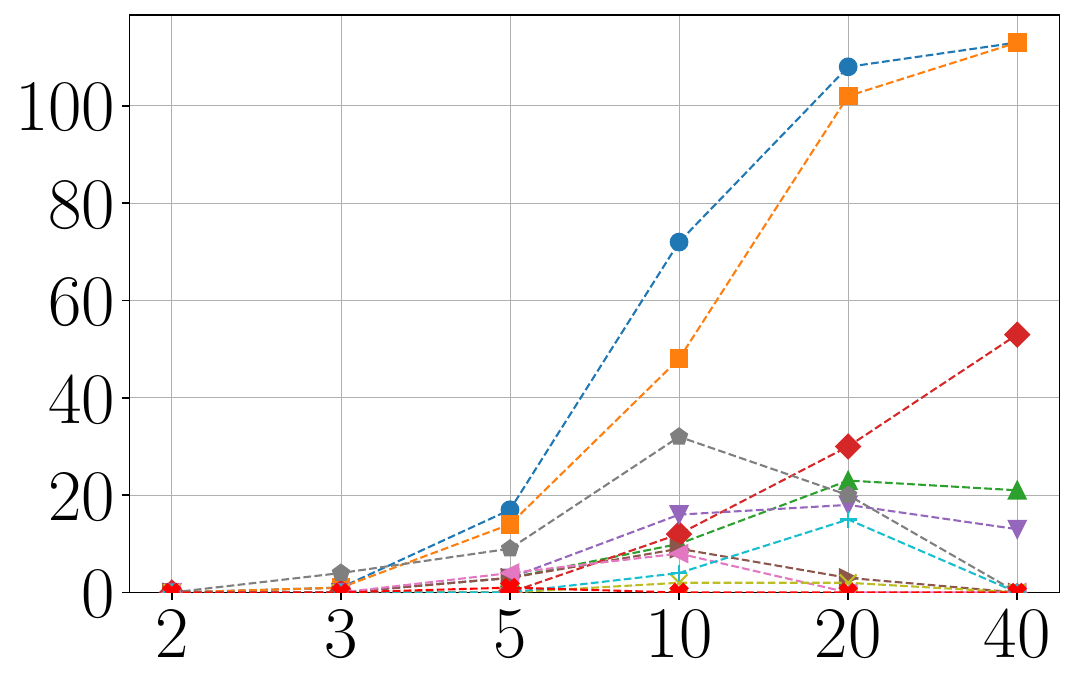}}
\subfloat[$\lambda=128 \lambda_{\text{def}}$]{\includegraphics[width=\wir\textwidth]{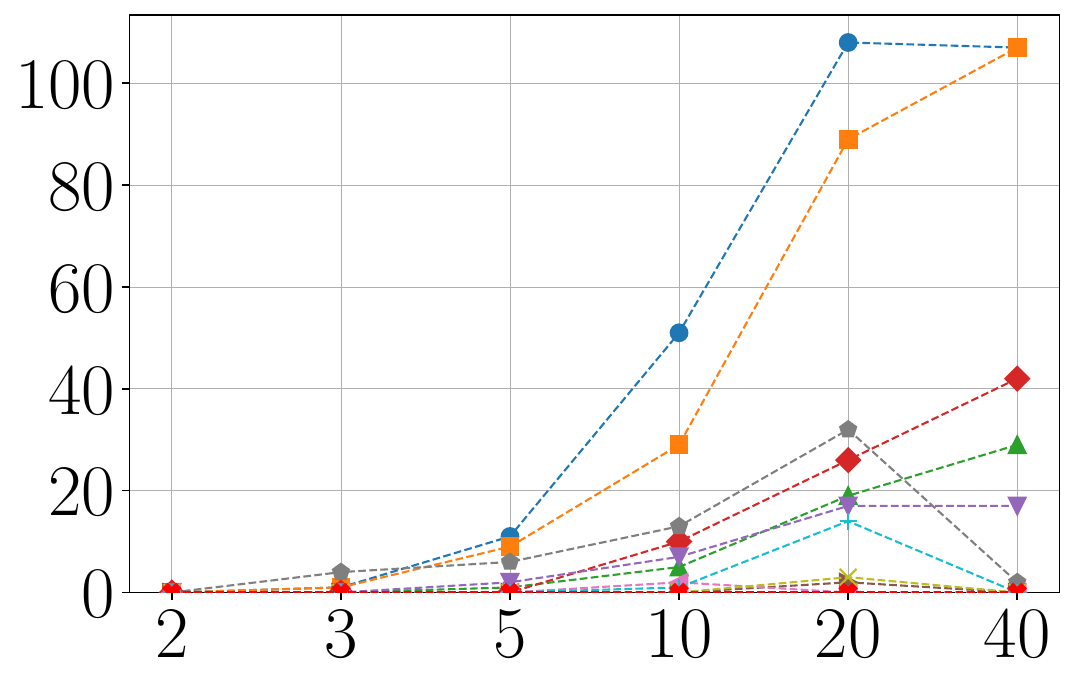}}
\subfloat[$\lambda=256 \lambda_{\text{def}}$]{\includegraphics[width=\wir\textwidth]{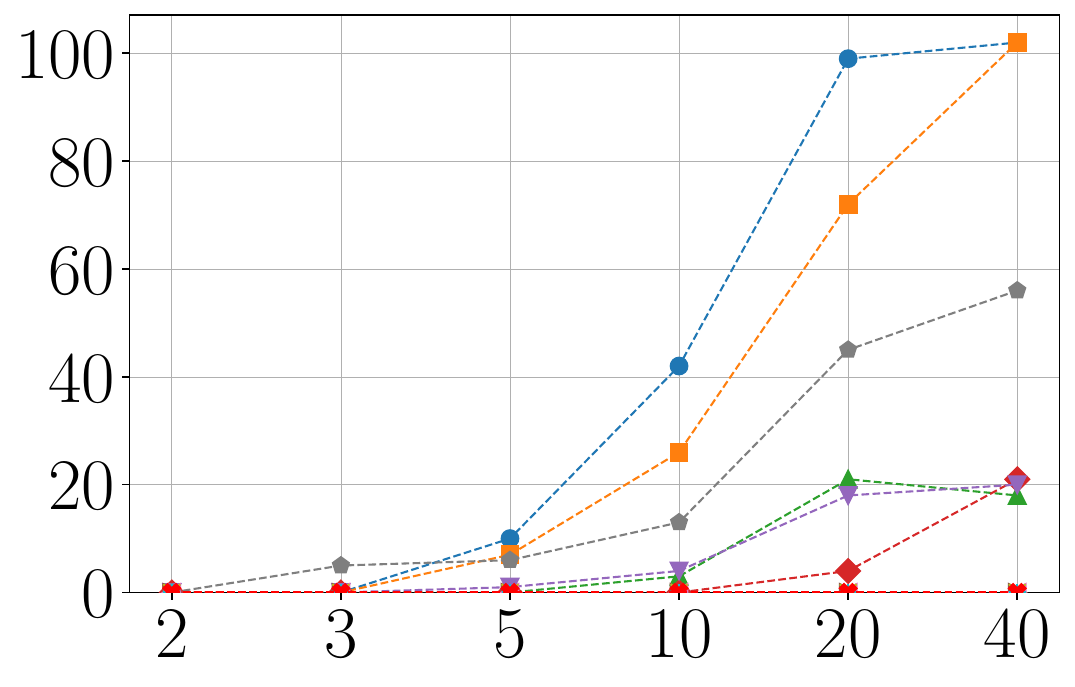}}
\caption{Number of times each stopping criterion stops the search before $\texttt{FE}^*$. The \textbf{horizontal axis} shows the dimension, \(n \in \{2, 3, 5, 10, 20, 40\}\), and the \textbf{vertical axis} shows the frequency.}
\label{fig:ex_count}
\end{figure}



\subsubsection*{Summary and discussion}

Our results show that \texttt{portfolio} and the three fitness-based criteria (\texttt{tolfun}, \texttt{tolfunhist}, and \texttt{tolflatfitness}) perform well in most cases.
However, the stopping accuracy of \texttt{tolfun} degrades as $\lambda$ increases, whereas that of \texttt{tolflatfitness} degrades as $n$ increases.
As demonstrated in \pref{sec:ftc}, \texttt{tolflatfitness} and \texttt{tolfun} are the stopping criteria that are most frequently triggered first.
However, they achieve worse stopping accuracy than \texttt{tolfunhist}, especially when $\lambda$ and $n$ are large.

Although a portfolio of multiple stopping criteria has traditionally been used in CMA-ES, this practice has not previously been validated either theoretically or empirically.
This work provides the first empirical evidence supporting the use of multiple stopping criteria.

\subsection{Frequency of stopping before $\texttt{FE}^*$}
\label{sec:results_early}

\pref{fig:ex_count} shows the number of times each stopping criterion stops the search before $\texttt{FE}^*$ on the $360$ function instances.
Such cases indicate that CMA-ES is stopped even though further search could still yield a better solution.
To avoid missing such potential improvements, stopping criteria should be designed so as not to trigger before $\texttt{FE}^*$.
Thus, in \pref{fig:ex_count}, a smaller y-axis value indicates better performance with respect to this property.

As shown in \pref{fig:ex_count}(a), when $\lambda=1 \lambda_{\text{def}}$, all stopping criteria are triggered before $\texttt{FE}^*$ in at least one of the 360 runs.
The number of such cases increases as the dimension $n$ increases.
Clearly, the \texttt{tolfun} and \texttt{tolfunhist} criteria stop the search before $\texttt{FE}^*$ most frequently in high dimensions.
As shown in \pref{fig:ex_count}(b)--(i), the same trend is observed for the \texttt{tolfun} and \texttt{tolfunhist} criteria as $\lambda$ increases.
Recall that, as observed in \pref{sec:results_pose}, some stopping criteria (e.g., \texttt{tolfacupx}) are triggered only rarely, especially for large values of $\lambda$.
Although this occurs less frequently than for \texttt{tolfun} and \texttt{tolfunhist}, \texttt{tolstagnation} also stops the search before $\texttt{FE}^*$ as $n$ increases.
As shown in \pref{fig:ex_count}, \texttt{tolupsigma} is frequently triggered before $\texttt{FE}^*$ for particular combinations of $\lambda$ and $n$, such as $\lambda = 64\lambda_{\text{def}}$ and $n = 10$ in \pref{fig:ex_count}(g).


\subsubsection*{Summary and discussion}

Our results show that the \texttt{tolfun} and \texttt{tolfunhist} criteria are triggered before $\texttt{FE}^*$ most frequently in high dimensions.
In~\cite{MartyHASH24}, Marty et al. proposed a variant of CMA-ES for black-box mixed-integer optimization. 
They briefly noted that the \texttt{tolfun} criterion tends to stop the search too early on the mixed-integer BBOB problem set (\texttt{bbob-mixint})~\cite{TusarBH19}, particularly in high dimensions.
Although the type of optimization problem is different (\texttt{bbob} or \texttt{bbob-mixint}), we quantitatively validated their observation.
We used the default hyperparameter values for the stopping criteria.
The issue with \texttt{tolfun} and \texttt{tolfunhist} may be addressed by tuning their hyperparameter values.

%% file: conclusion.tex
\section{Conclusion}
\label{sec:conclusion}

This paper analyzed 11 stopping criteria in CMA-ES, excluding \texttt{tolfunrel}.
First, \pref{sec:review_sc} reviewed the 12 stopping criteria implemented in \texttt{pycma}.
Then, \pref{sec:ftc} investigated which stopping criterion is triggered first most frequently in CMA-ES.
\pref{sec:results_pose} examined the stopping accuracy of each stopping criterion in CMA-ES.
Finally, \pref{sec:results_early} investigated how frequently the stopping criteria stop the search before CMA-ES reaches complete stagnation.
The main findings of Sections \ref{sec:ftc}--\ref{sec:results_early} are summarized at the end of each section.

Our findings are helpful for developing more accurate stopping criteria for CMA-ES, which can directly improve the performance of restart variants such as IPOP-CMA-ES and BIPOP-CMA-ES.
For example, except for \texttt{tolupsigma}, we observed that most convergence-based and divergence-based criteria are triggered too late.
In other words, these criteria are too conservative under the default hyperparameter settings.\footnote{These criteria are designed not only to detect stagnation but also to prevent crashes. This may explain why they were triggered too late in this study.}
Their performance may be improved by adjusting the hyperparameters to make them more aggressive.
In contrast, we observed that the \texttt{tolfun} and \texttt{tolfunhist} criteria are overly aggressive.
Their performance may benefit from more conservative hyperparameter settings.
We believe that automatic algorithm configuration~\cite{LopezIbanez16} can effectively identify a good hyperparameter configuration by using the POSE value in \pref{eq:pose} as the objective function.
It would also be interesting to tune all hyperparameters of the 11 stopping criteria simultaneously so that their portfolio achieves the best performance.

While this paper focused on an empirical analysis of the stopping criteria, their theoretical analysis remains future work.
An investigation into the invariance properties of the stopping criteria may be worthwhile.
As implicitly pointed out in \cite{CuccuGG11}, some stopping criteria (e.g., \texttt{tolfun}) are not invariant.
For example, \texttt{tolfun} is not generally invariant under order-preserving transformations of the objective function values, because it depends on absolute differences in objective values.
Although CMA-ES has desirable invariance properties, restart CMA-ES variants using \texttt{tolfun} are not invariant under objective value transformations.
Our analysis was based on aggregated results over the 24 \texttt{bbob} functions and 15 instances for each function.
A more detailed analysis of the behavior of the stopping criteria on individual function instances remains an important direction for future research.
Although this paper focused only on CMA-ES, future work should investigate stopping criteria in other evolutionary algorithms, such as differential evolution~\cite{StornP97}.
